\documentclass{article}
\usepackage{fancyhdr}

\usepackage{report}

\usepackage{array}
\usepackage{booktabs}

\usepackage{tabularx}
\usepackage{xcolor}
\usepackage{colortbl}
\usepackage[most]{tcolorbox}

\newcolumntype{Y}{>{\raggedright\arraybackslash}X}
\renewcommand{\arraystretch}{1.18}
\newcolumntype{C}{>{\centering\arraybackslash}X} 

\usepackage{microtype}
\usepackage{graphicx}
\usepackage{subfigure}
\usepackage{booktabs} 

\usepackage{hyperref}

\usepackage{pifont}
\usepackage{soul}
\RequirePackage{natbib}
\usepackage{makecell}
\usepackage{graphicx} 
\usepackage[utf8]{inputenc} 
\usepackage{CJKutf8}
\usepackage[T1]{fontenc}

\usepackage{hyperref}       
\usepackage{url}            
\usepackage{booktabs}       
\usepackage{amsfonts}       
\usepackage{fancyhdr}       

\usepackage{nicefrac}       
\usepackage{microtype}      
\usepackage{graphicx}
\usepackage{pifont}
\usepackage{multirow}
\usepackage{CJKutf8}
\definecolor{darkmagenta}{rgb}{0.56, 0.0, 1.0}
\definecolor{softyellow}{rgb}{1.0, 0.92, 0.3} 
\definecolor{LightAquamarine}{rgb}{0.75, 1.0, 0.8} 
\definecolor{FireBrick}{RGB}{178,34,34}
\definecolor{MediumPurple}{RGB}{147,112,219}

\definecolor{uclablue}{rgb}{0.15, 0.45, 0.68}
\hypersetup{
    breaklinks,
    colorlinks=true,
    citecolor={darkmagenta},
    linkcolor={uclablue},
    urlcolor={uclablue}
}
\usepackage{wrapfig}
\usepackage{float}
\usepackage{subcaption}
\usepackage{placeins}
\usepackage{lipsum} 
\usepackage{tcolorbox}
\usepackage{amsmath}
\usepackage{amssymb}
\usepackage{tcolorbox}
\usepackage{fancyvrb}
\usepackage{verbatim}

\usepackage{utfsym}
\usepackage{xspace}
\usepackage{enumitem}
\usepackage{multirow} 
\tcbuselibrary{breakable}
\usepackage{enumitem}
\usepackage{colortbl}
\usepackage{fancyhdr}
\usepackage{tabularx}
\usepackage{tcolorbox}
\usepackage{transparent}

\definecolor{njuPurple}{RGB}{220,205,230}     
\definecolor{njuPurpleLight}{RGB}{250,245,252}   

\newtcolorbox{abstractbox}{
    colback=njuPurpleLight,   
    colframe=njuPurple,       
    boxrule=1pt,              
    arc=4mm,                  
    left=8pt,                 
    right=8pt,                
    top=8pt,                  
    bottom=8pt,               
    opacityback=0.95
}

\usepackage[table]{xcolor}
  \usepackage{pifont}
  \usepackage{booktabs}
  \usepackage{multirow}
  \usepackage{makecell}
  \usepackage{graphicx}
  \definecolor{groupcolor}{HTML}{FBE5D6}   
  \definecolor{ourscolor}{HTML}{E2F0D9}    
  \definecolor{headercolor}{HTML}{F2F2F2}  
  \newcommand{\cmk}{\textcolor{green!55!black}{\ding{51}}}
  \newcommand{\xmk}{\textcolor{red!70!black}{\ding{55}}}

  \usepackage{tikz}

  \definecolor{vfblue}{HTML}{4472C4}
  \definecolor{vfbluefill}{HTML}{D9E2F3}
  \definecolor{prgreen}{HTML}{548235}
  \definecolor{prgreenfill}{HTML}{E2EFDA}
  \definecolor{clpurple}{HTML}{7030A0}
  \definecolor{clpurplefill}{HTML}{E4D7EE}
  \definecolor{ourscolor}{HTML}{E2F0D9}

  \newcommand{\pillbase}[3]{%
    \tikz[baseline=(b.base)]\node[draw=#1, fill=#2, rounded corners=3pt,
          inner xsep=4pt, inner ysep=1.5pt, font=\scriptsize\bfseries,
          text=#1] (b) {#3};%
  }
  \newcommand{\pillVF}{\pillbase{vfblue}{vfbluefill}{VF}}
  \newcommand{\pillPR}{\pillbase{prgreen}{prgreenfill}{PR}}
  \newcommand{\pillCL}{\pillbase{clpurple}{clpurplefill}{CL}}
\usepackage{array}
\usepackage{booktabs}
\usepackage{tabularx}
\usepackage{makecell}
\usepackage{multirow}
\usepackage{graphicx}
\usepackage{wrapfig}
\usepackage{float}
\usepackage{subcaption}
\usepackage{placeins}

\usepackage{amsmath}
\usepackage{amssymb}
\usepackage{amsfonts}
\usepackage{nicefrac}
\usepackage{pifont}
\usepackage{utfsym}
\usepackage{xspace}

\usepackage{xurl}
\usepackage{hyperref}

\usepackage{enumitem}
\usepackage{soul}
\usepackage{xcolor}
\usepackage{colortbl}
\usepackage[most]{tcolorbox}
\usepackage{fancyvrb}
\usepackage{verbatim}
\usepackage{transparent}
 \usepackage{listings}
  \usepackage{xcolor}

  \definecolor{boxgray}{gray}{0.95}

  \lstdefinestyle{promptstyle}{
      basicstyle=\ttfamily\scriptsize,
      breaklines=true,
      breakatwhitespace=false,
      showstringspaces=false,
      columns=fullflexible,
      frame=single,
      framesep=4pt,
      rulecolor=\color{black!40},
      backgroundcolor=\color{boxgray},
      xleftmargin=4pt,
      xrightmargin=4pt,
      aboveskip=6pt,
      belowskip=6pt,
      upquote=true,
      literate={—}{{---}}1 {→}{{$\rightarrow$}}1 {↔}{{$\leftrightarrow$}}1 {↑}{{$\uparrow$}}1
  {↓}{{$\downarrow$}}1 {–}{{--}}1 {“}{{``}}1 {”}{{''}}1 {’}{{'}}1 {‘}{{'}}1
  }

\newtcbox{\tagbox}[1][]{%
  on line,
  boxrule=0.35pt,
  arc=2pt,
  boxsep=0pt,
  left=4pt,right=4pt,top=1.2pt,bottom=1.2pt,
  colframe=black!18,
  coltext=black,
  fontupper=\scriptsize\sffamily,
  #1
}

\newtcolorbox{casebox}[2][]{%
  enhanced,
  breakable,
  colback=gray!2,
  colframe=black!18,
  boxrule=0.4pt,
  arc=2pt,
  left=4pt,right=4pt,top=2pt,bottom=2pt,
  fonttitle=\bfseries\small,
  title={#2},
  #1
}

\usepackage[most]{tcolorbox}

\tcbset{
  mainbox/.style={
    enhanced,
    colback=white,
    colframe=black!20,
    boxrule=0.5pt,
    arc=1mm
  },
  titlebox/.style={
    enhanced,
    colback=black!60,
    colframe=black!60,
    coltext=white,
    boxrule=0pt,
    arc=1mm,
    left=1mm,
    right=1mm,
    top=0.5mm,
    bottom=0.5mm,
    fontupper=\bfseries
  }
}

\title{CoVEBench: Can Video Editing Models Handle Complex Instructions?}

\author{
\textbf{Jiangtao Wu}$^{1,*}$\quad
\textbf{Jiaming Wang}$^{1,*}$\quad
\textbf{Yiwen He}$^{1,*}$\quad
\textbf{Yuanxing Zhang}$^{2}$\quad
\textbf{Shihao Li}$^{1}$\\
\textbf{Dunyuan Liu}$^{1}$\quad
\textbf{Xuedong Zhao}$^{1}$\quad
\textbf{Jialu Chen}$^{2}$\quad
\textbf{Zekun Moore Wang}$^{2}$\quad
\textbf{Jiaheng Liu}$^{1,\dagger}$\\
\vspace{4mm}
{\normalsize $^1$ NJU-LINK Team, Nanjing University} \quad
{\normalsize $^2$ Kling Team, Kuaishou Technology}\\
\vspace{2mm}
\texttt{jiangtaowu@smail.nju.edu.cn}
\quad
\texttt{liujiaheng@nju.edu.cn}
}

\let\oldthefootnote\thefootnote

\let\thefootnote\relax\footnotetext{*~Equal Contribution. ~~$^\dagger$~Corresponding Author.}
\let\thefootnote\oldthefootnote
\begin{document}
\maketitle
\begin{abstractbox}
\begin{center}
\textbf{\Large Abstract}
\end{center}
\noindent
While recent text-guided video editing models excel at elementary tasks (e.g., style transfer, object insertion), real-world user requests are highly compositional. A single prompt often demands multiple coupled edits, such as modifying subjects, actions, and camera views, while strictly preserving unrelated spatiotemporal content. Existing benchmarks, heavily constrained by isolated edits and coarse global metrics, fail to diagnose how models handle such complex workflows. To address this gap, we introduce \textbf{CoVEBench}~\footnote{\url{https://github.com/NJU-LINK/CoVEBench}} \footnote{\url{https://huggingface.co/datasets/NJU-LINK/CoVEBench}}, a compositional video editing benchmark comprising 416 curated source videos, 626 multi-point editing instructions, and 9,990 fine-grained checklist items. Covering diverse editing dimensions, CoVEBench evaluates models via MLLM-judged instruction compliance and video fidelity, alongside automated metrics for video quality. Extensive experiments reveal that compositional editing remains a profound challenge: current models frequently omit edits, violate preservation constraints, or introduce artifacts when handling multiple operations simultaneously. CoVEBench provides a challenging, diagnostic testbed to advance video editing toward realistic user workflows.

\end{abstractbox}

\section{Introduction}
Instruction-guided video editing models (e.g., Seedance 2.0 \citep{seedance2026seedance20advancingvideo}) have advanced rapidly, significantly accelerating video creation with striking capabilities in tasks like style transfer~\citep{Yang2023RerenderAV} and object replacement~\citep{Gu2023VideoSwapCV}. However, current evaluations severely lag behind. By predominantly focusing on simple, isolated edits, existing benchmarks have become too easy for today's advanced proprietary models, creating a stark disconnect from real-world workflows~\citep{Brooks2022InstructPix2PixLT, huang2023vbench}. In practice, user requests are inherently compositional. Creators typically demand simultaneous multi-point edits within a single prompt, such as modifying subjects, adjusting camera motion, and adding objects while preserving the background. Compared with single-point editing, such compositional editing tests more fundamental model capabilities: whether the model can understand the relationships among multiple editing goals, correctly combine several operations without mutual interference, and strictly preserve irrelevant content while modifying the target regions. Handling such complexity requires models to coordinate multiple atomic operations under shared spatiotemporal constraints~\citep{Geyer2023TokenFlowCD}. Therefore, to align model evaluation with real-world creation demands, there is an urgent need for a benchmark dedicated to complex, compositional video editing.

\begin{figure*}[tb!]
  \centering
  \includegraphics[width=1\textwidth]{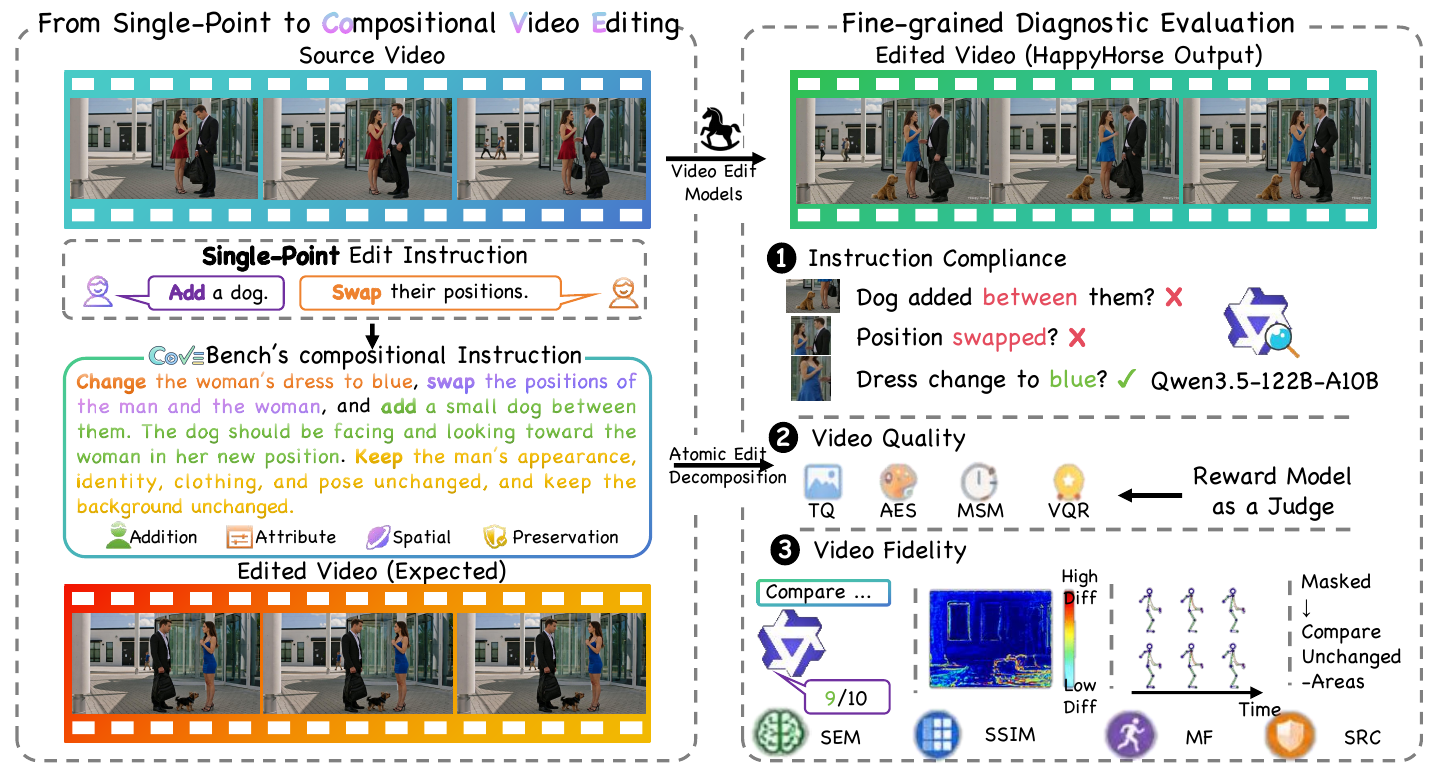}
  \caption{Video editing is moving towards complex instructions. CoVEBench provides evaluation for compositional video editing with fine-grained diagnostics across instruction compliance, video quality, and video fidelity.}
  \label{fig:overview}
\end{figure*}

Beyond the lack of complex instructions, current evaluation protocols are fundamentally inadequate for measuring such compositional workflows. In realistic scenarios, multiple editing operations inevitably interact or interfere with one another. However, existing benchmarks rely heavily on coarse, global metrics (e.g., CLIP scores~\citep{Radford2021LearningTV,Gal2021StyleGANNADA}), which are incapable of assessing these complex dynamics~\citep{Hessel2021CLIPScoreAR}. Such global measurements cannot diagnose specific failures~\citep{openai2023gpt4}: they fail to reveal if a model successfully executed one edit but omitted another, generated a visually plausible but physically illogical modification, or unintentionally altered irrelevant scene structures. Therefore, a realistic benchmark must go beyond merely providing compositional prompts; it must establish fine-grained diagnostic metrics to rigorously evaluate execution accuracy, modification quality, and semantic preservation.

To address this gap, we introduce \textbf{CoVEBench}, a benchmark designed for compositional instruction-guided video editing. CoVEBench contains 416 curated source videos and 626 multi-point editing instructions, with each instruction specifying approximately 3 atomic edit operations on average. These operations cover a comprehensive range of fundamental editing dimensions, such as subject editing, background modification, camera work, style transfer, motion change, positional relationship adjustment, and special effects. To enable interpretable evaluation, we decompose these complex instructions into 9,990 verifiable checklist items. Assessed via state-of-the-art Multimodal Large Language Models (MLLMs), these items systematically diagnose whether each edit is successfully executed, whether the modified region remains visually and logically plausible, and whether unrelated content is properly preserved~\citep{Liu2023VisualIT}. In addition to this checklist-based compliance, CoVEBench incorporates automatic metrics to comprehensively evaluate video quality, motion consistency, and structural preservation. Together, CoVEBench provides a challenging and diagnostic testbed for evaluating whether current models can move beyond isolated edits and handle the compositional workflows demanded by real users.

In summary, our contributions are threefold:
\begin{itemize}
\item \textbf{Novel Task and Benchmark:} We propose a task of compositional video editing for complex scenarios, targeting realistic multi-point editing workflows. To facilitate this, we introduce \textbf{CoVEBench}, comprising 416 curated source videos and 626 complex editing instructions with an average of approximately 3 atomic edit operations per instruction.
\item \textbf{Reliable Evaluation System:} We establish a reliable and comprehensive evaluation framework. By constructing 9,990 fine-grained checklist items alongside automatic video quality metrics, our system provides a diagnostic assessment of execution accuracy, modification realism, and semantic preservation.
\item \textbf{Extensive Benchmarking and Insights:} We conduct extensive evaluation on a wide range of leading open-source and proprietary video editing models. Our results reveal significant performance gaps between current models and the demands of complex compositional editing, providing insights for future research.
\end{itemize}

\section{Related Work}
\paragraph{Evolution of Video Editing Models.}
Text-driven video editing has evolved from prompt-based adaptation to instruction-following paradigms. Early approaches extended pre-trained text-to-image models with temporal mechanisms (e.g., cross-frame attention~\citep{Wu2022TuneAVideoOT}, optical-flow~\citep{Cong2023FLATTENOF}) to maintain frame consistency. However, reliance on per-video optimization or rigid source-target prompt pairs limited their flexibility. Consequently, recent methods~\citep{Brooks2022InstructPix2PixLT, Jiang_2025_ICCV, Wu2025InsViE1MEI,cao2026t2avcompass} have shifted towards instruction-driven systems, leveraging large-scale triplets and unified frameworks to enhance open-ended generalization. Despite these advances, existing models are mostly evaluated on simple, isolated edits. Their capability to execute compositional instructions---which demand multiple coupled edits while preserving irrelevant source content---remains insufficiently explored.

\paragraph{Video Editing Benchmarks.}
 \begin{table*}[t]
  \renewcommand{\arraystretch}{1.25}
  \setlength{\tabcolsep}{3.4pt}
  \resizebox{\linewidth}{!}{
  \begin{tabular}{l cc cccccc c}
  \toprule
  \multirow{2}{*}{\textbf{Benchmark}} &
  \multicolumn{2}{c}{\textbf{Collection}} &
  \multicolumn{6}{c}{\textbf{Prompt}} &
  \multirow[c]{2}{*}{
\makecell[c]{\textbf{Evaluation}\\\textbf{Dimension}\\\textbf{\& Metrics}}
}\\
  \cmidrule(lr){2-3} \cmidrule(lr){4-9}
  & \makecell[c]{Source\\Videos}
  & \makecell[c]{Prompt\\Count}
  & \makecell[c]{Category\\Count}
  & \makecell[c]{Avg.\\Words}
  & \makecell[c]{Compositional\\Editing}
  & \makecell[c]{Motion\\Editing}
  & \makecell[c]{Camera\\Editing}
  & \makecell[c]{Visual\\Effect}
  & \\
  \midrule

  EditBoard~\citep{chen2025editboardcomprehensiveevaluationbenchmark}
  & 40   & 80
  & 4   & 7.4
  & \cmk & \xmk & \xmk & \xmk
  & \pillVF \\

  VE-Bench~\citep{sun2024vebenchsubjectivealignedbenchmarksuite}
  & 169  & 197
  & 3   & 8.2
  & \xmk & \xmk & \xmk & \xmk
  & \pillVF \\

  FiVE~\citep{Li_2025_ICCV}
  & 100  & 420
  & 6   & 7.8
  & \xmk & \xmk & \xmk & \xmk
  & \pillVF~\pillCL \\


  TDVE-Assessor~\citep{Wang2025TDVEAssessorBA}
  & 180  & 340
  & 8   & {28.5}
  & \xmk & \cmk & \xmk & \xmk
  & \pillVF \\

  VEditBench~\citep{wu2025veditbench}
  & 420  & 2520
  & 6   & 7.7
  & \xmk & \cmk & \xmk & \xmk
  & \pillVF \\

  IVEBench~\citep{chen2026ivebenchmodernbenchmarksuite}
  & 600  & 600
  & 12  & 8.2
  & \xmk & \cmk & \cmk & \cmk
  & \pillVF \\

  \midrule
  \rowcolor{ourscolor}
  \textbf{CoVEBench (Ours)}
  & \textbf{416} & \textbf{626}
  & \textbf{19} & \textbf{44.9}
  & \cmk & \cmk & \cmk & \cmk
  & \pillVF~\pillPR~\pillCL \\

  \bottomrule
  \end{tabular}
  }
  \centering
    \caption[Comparison of representative video editing benchmarks.]{%
Comparison of representative video editing benchmarks.
\protect\pillVF\,Video Fidelity,
\protect\pillPR\,Physical Realism, and
\protect\pillCL\,Objective QA evaluated by MLLM-based judging.
The average prompt length of TDVE-Assessor is computed from edited-video captions rather than edit instructions. More comparisons are provided in Appendix~\ref{app:com}.
}
  \label{tab:comparison}
  \end{table*}

While existing benchmarks have driven progress, a systematic comparison (Table~\ref{tab:comparison}) reveals three limitations in realistic workflows:
(1) Simplistic prompts: They focus on isolated edits~\citep{chen2026ivebenchmodernbenchmarksuite,Jiang_2025_ICCV,Wang2025TDVEAssessorBA,Li_2025_ICCV}. Even when compositional instructions exist, they are rare and confined to trivial attribute modifications~\citep{chen2025editboardcomprehensiveevaluationbenchmark}.
(2) Narrow scope: They noticeably lack structural and dynamic operations, such as camera and subject motion edits~\citep{sun2024vebenchsubjectivealignedbenchmarksuite}.
(3) Coarse evaluation: Relying on holistic scores~\citep{wu2025veditbench} or general reward models, they fail to measure fine-grained adherence to complex prompts or identify physical law violations.
To address these gaps, CoVEBench emphasizes compositional editing, broadens operation categories, and employs fine-grained checklists to accurately evaluate multi-point instruction adherence and physical realism.

\section{CoVEBench}

Fig.~\ref{fig:data_statistics} summarizes the statistics of CoVEBench. Our hierarchical taxonomy ensures diverse coverage of editing dimensions (Fig.~\ref{fig:data_statistics}a), while token and keyword distributions confirm broad semantic richness (Fig.~\ref{fig:data_statistics}b,c). Quantitatively, the finalized instructions exhibit an extremely low TF-IDF cosine similarity across 195,625 sampled pairs (mean: 0.0168, median: 0.0106). CoVEBench comprises \textbf{416} curated videos with varied durations and resolutions (Fig.~\ref{fig:data_statistics}e,f). Notably, our prompts emphasize complex and multi-point edits rather than isolated ones (Fig.~\ref{fig:data_statistics}d), enabling rigorous evaluation of compositional editing capabilities.

\subsection{Data Construction Pipeline}
\begin{figure*}[tb!]
  \centering
  \includegraphics[width=1\textwidth]{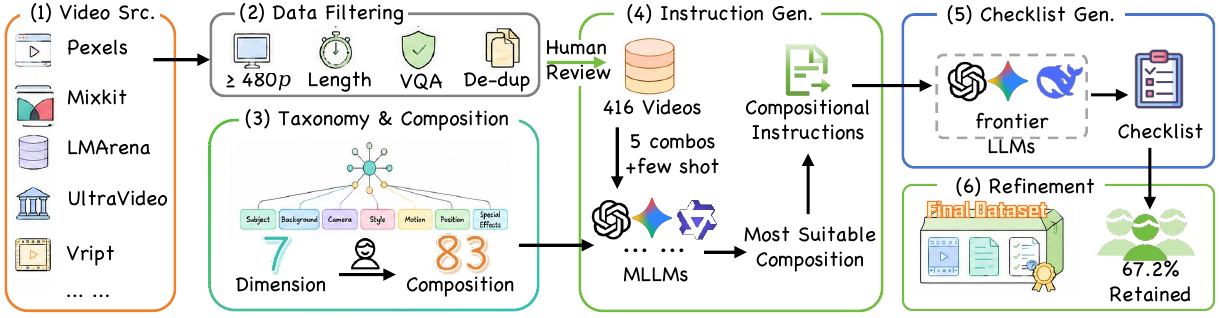}
  \caption{Data curation pipeline of CoVEBench.}
  \label{fig:data_pipeline}
\end{figure*}
As illustrated in Fig.~\ref{fig:data_pipeline}, CoVEBench is constructed through a three-stage pipeline: source video collection and filtering, editing instruction generation, and checklist generation with refinement.

\paragraph{Video Collection.}
To construct a diverse source video pool spanning various subjects and scenes, we integrate five complementary sources: stock platforms (Pexels, Mixkit) and academic datasets (Vript~\citep{li2025ifvidcapvideocaptionmodels, yang2024vriptvideoworththousands}, UltraVideo~\citep{xue2025ultravideohighqualityuhdvideo}, ViDiC~\citep{wu2026vidicvideodifferencecaptioning} and LMArena). We then apply strict filtering criteria, including resolution $\ge$480p, duration within 3--21s, visual quality screening, and cross-pool near-duplicate removal. After a final human review for editability and overall quality, we select 416 videos.

\paragraph{Editing Instruction Generation.}
To ensure instruction diversity and cover real-world editing scenarios, we establish a structured taxonomy comprising seven practical dimensions (Fig.~\ref{fig:data_statistics}a; detailed in Appendix~\ref{app:category}): \textbf{Subject}, \textbf{Background}, \textbf{Camera}, \textbf{Style}, \textbf{Motion}, \textbf{Position}, and \textbf{Special Effects}. Based on this taxonomy, we manually formulate 83 distinct category combinations that closely reflect realistic editing workflows. We then distribute the source videos across a diverse pool of MLLMs, including GPT-5~\citep{openai2025gpt5}, Gemini-3.1-Pro~\citep{googledeepmind2026gemini31pro}, Qwen3-VL-plus~\citep{qwen3}, and Doubao-Seed-1.8~\citep{seed2025}. For each assigned video, the specific model is prompted with a dynamically rotated subset of 5 combinations and varying few-shot examples. The model selects the most suitable combination and generates a tailored editing instruction, followed by a manual review to remove inappropriate or repetitive outputs. 

\paragraph{Checklist Generation.}
To systematically evaluate whether the specified edits are successfully executed, we employ advanced LLMs, including Gemini-3-Flash~\citep{googledeepmind2025gemini3flash}, GPT-5, and DeepSeek-V4-Pro~\citep{deepseekai2026deepseekv4}, to synthesize checklist questions. Given the editing instruction and a detailed text description of the source video, the models extract distinct editing points and reorganize them into fine-grained, verifiable questions. After rigorous manual filtering, we retain approximately 67.2\% of the initial outputs, yielding a reliable checklist framework for assessing execution accuracy, modification quality, and semantic preservation (see Appendix~\ref{app:test}).

\subsection{Data Statistics}
\begin{figure*}[tb!] 
  \centering
  \includegraphics[width=1\textwidth]{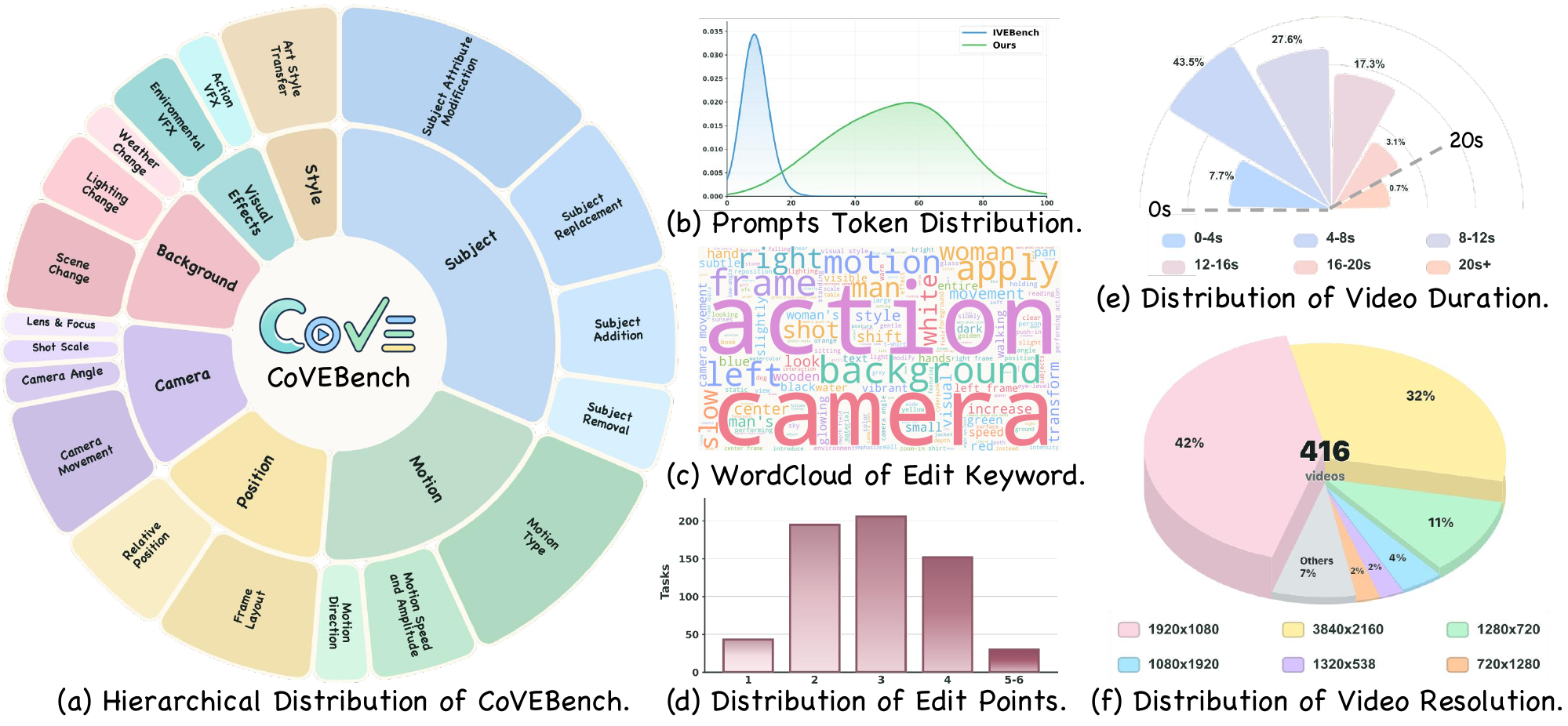}
  \caption{Data statistics of CoVEBench, showing broad coverage across edit types and video properties. Detailed video topic distributions are provided in Appendix~\ref{sec:topic-distribution}.}
  \label{fig:data_statistics}
\end{figure*}
Fig.~\ref{fig:data_statistics} summarizes the statistics of CoVEBench. Our hierarchical taxonomy ensures diverse coverage of editing dimensions (Fig.~\ref{fig:data_statistics}a), while token and keyword distributions confirm broad semantic richness (Fig.~\ref{fig:data_statistics}b,c). Quantitatively, the finalized instructions exhibit an extremely low TF-IDF cosine similarity across 195,625 sampled pairs (mean: 0.0168, median: 0.0106). CoVEBench comprises \textbf{416} curated videos with varied durations and resolutions (Fig.~\ref{fig:data_statistics}e,f). Notably, our prompts emphasize complex and multi-point edits rather than isolated ones (Fig.~\ref{fig:data_statistics}d), enabling rigorous evaluation of compositional editing capabilities.

\subsection{Evaluation Methodology}

\begin{table*}[tb!]

  \resizebox{\textwidth}{!}{%
  \renewcommand{\arraystretch}{1.2}
  \begin{tabular}{@{} l l l l @{}}
  \toprule
  \textbf{Dimension} & \textbf{Metric (Abbr.)} & \textbf{Method / Tool} & \textbf{Evaluation Focus}
  \\
  \midrule

  \multirow{3}{*}{\makecell[l]{\textbf{Instruction}\\ \textbf{Compliance}}}
  & \cellcolor{ourscolor}\textbf{Union Accuracy (UAS)}
  & \cellcolor{ourscolor}\textbf{MLLM + Checklist}
  & \cellcolor{ourscolor}\textbf{Success only if both instruction and realism queries are correct}
  \\
  & Instruction Following (IFS) & MLLM + Checklist & Measures edit execution regardless of editing
  quality \\
  & Video Realism (VRS) & MLLM + Checklist & Assesses visual naturalness and overall editing quality
   \\
  \midrule

  \multirow{4}{*}{\makecell[l]{\textbf{Video}\\ \textbf{Quality}}}
  & \cellcolor{ourscolor}\textbf{Comprehensive Quality (VQR)}
  & \cellcolor{ourscolor}\textbf{VisualQuality-R1}
  & \cellcolor{ourscolor}\textbf{Holistic visual quality assessment} \\
  & Aesthetics (AES) & Aesthetic Predictor v2.5 & Frame \& keyframe-level visual appeal \\
  & Motion Smoothness (MSM) & Optical-flow fields & Temporal motion stability \\
  & Technical Quality (TQ) & DOVER++ & Perceived technical distortions \\
  \midrule

  \multirow{4}{*}{\makecell[l]{\textbf{Video}\\ \textbf{Fidelity}}}
  & \cellcolor{ourscolor}\textbf{Semantic Consistency (SEM)}
  & \cellcolor{ourscolor}\textbf{MLLM + Checklist}
  & \cellcolor{ourscolor}\textbf{Context-aware preservation of unedited semantics} \\
  & Structural Fidelity (SSIM) & SSIM & Pixel-level layout and structure preservation \\
  & Motion Fidelity (MF) & CoTracker & Temporal motion trajectory similarity \\
  & Static Region Consistency (SRC) & SAM2 + DINOv2 & Mask-based feature comparison for unchanged
  areas \\

  \bottomrule
  \end{tabular}
  }
    \centering
  \caption{Evaluation matrix of CoVEBench. Metrics are organized into three dimensions. The green-highlighted metrics (UAS, VQR, and SEM) are the primary holistic indicators for Instruction Compliance, Video Quality, and Video Fidelity, respectively, aggregating the information of all subordinate metrics within each dimension.}
  \label{tab:evaluation-matrix}
  \end{table*}

As summarized in Table~\ref{tab:evaluation-matrix}, we evaluate editing models across three complementary dimensions: \textbf{Instruction Compliance} (adherence to prompts and the realism of the executed edits), \textbf{Video Quality} (aesthetic appeal, temporal stability, low-level visual quality, and absence of AI artifacts), and \textbf{Video Fidelity} (preservation of unedited source content). We provide a brief introduction to our automatic and MLLM-based metrics below; please refer to Appendix~\ref{app:evaluation-details} for full details.

\paragraph{Instruction Compliance.} 
We evaluate compliance through a fine-grained MLLM checklist comprising two types of verifiable questions. 
(1) \textbf{Multiple-Choice Questions (MCQs)} formulate the pre-edit and post-edit states as distinct options, where the correct choice strictly matches the expected outcome specified in the instruction. 
(2) \textbf{Yes/No Questions} assess edit execution and basic potential unnatural artifacts. Depending on whether referencing the source content is necessary, we dynamically employ Dual-Video (comparing source and edited videos) or Single-Video formats. 

To quantitatively measure this, we derive three distinct metrics from the checklist to clearly separate execution accuracy from visual quality. The Instruction Following Score (IFS) evaluates strictly whether the requested editing actions were performed, regardless of the editing quality, by averaging the accuracy of instruction-focused queries. Conversely, the Video Realism Score (VRS) specifically assesses the editing quality, measuring the visual naturalness and overall realism of the edited content. To provide a rigorous overall evaluation, we introduce Union Accuracy (UAS) as a strict combined metric, where a specific edit task scores 1 only if both its instruction-following and realism questions are answered correctly. For this automated evaluation, we utilize \textbf{Qwen3.5-122B-A10B}~\citep{qwen3.5} as the MLLM judge.

\paragraph{Video Quality Metrics.}
We assess the quality of edited videos across five dimensions:
(1) \textbf{Aesthetics}: Measured by Aesthetic Predictor v2.5~\citep{aesthetic_predictor_v25}, which averages the aesthetic scores derived from both 10 uniformly sampled frames and the extracted video keyframes.
(2) \textbf{Motion Smoothness}: Evaluated by measuring the consistency of optical-flow fields between adjacent frames to capture temporal stability and inter-frame coherence.
(3) \textbf{Low-level Quality}: Assessed using the DOVER++~\citep{wu2023dover} technical score to capture pixel-level degradations.
(4) \textbf{Comprehensive Quality}: Evaluated by VisualQuality-R1~\citep{NEURIPS2025_7f8f7bf2}, a unified visual reward model that provides a holistic assessment of the video.

\paragraph{Video Fidelity and Preservation Metrics.}
To evaluate how well the edited video preserves the unaltered aspects of the source video, we assess fidelity across three hierarchical levels:
(1) \textbf{Structural and Motion Fidelity}: We measure pixel-level structural preservation using SSIM~\citep{1284395} and temporal motion consistency utilizing CoTracker~\citep{11444268}.
(2) \textbf{Static Region Consistency}: We localize specific regions expected to remain unaltered using GroundingDINO~\cite{liu2023grounding} and refine them with SAM2 masks~\cite{ravi2024sam2segmentimages}. Consistency is then evaluated by computing the cosine similarity between DINOv2 embeddings~\cite{Oquab2023DINOv2LR} extracted from these masked areas.
(3) \textbf{Semantic Consistency}: We introduce a tailored checklist to independently evaluate elements that should remain unaltered. Using Qwen3.5-122B-A10B as the evaluator, we prompt the model with specific questions, the pre- and post-edit videos, alongside the original edit instruction to assign a semantic consistency score on a 1-10 scale.

\section{Experiment}

\subsection{Main Results}

We evaluate 10 popular models including InsV2V~\citep{ICLR2024_48a13e12}, VACE~\citep{Jiang_2025_ICCV}, Lucy Edit~\citep{decart2025lucyedit}, ICVE~\citep{liao2025context}, Ditto~\citep{bai2025scalinginstructionbasedvideoediting}, Reco~\citep{Zhang2025RegionConstraintIG}, OmniWeaving~\citep{pan2026omniweavingunifiedvideogeneration}, Kiwi~\citep{kiwiedit} HappyHorse1.0~\citep{happyhorse2026} and Wan2.7\citep{wan2025wanopenadvancedlargescale}. The main results are presented in Table~\ref{tab:main-results}, which lead to the following key observations:
\begin{table*}[t]
\centering

\resizebox{\textwidth}{!}{
\begin{tabular}{l|ccc|cccc|cccc}
\toprule
\multirow{2}{*}{\textbf{Model}} 
& \multicolumn{3}{c|}{\textbf{Instruction Compliance}} 
& \multicolumn{4}{c|}{\textbf{Video Quality}} 
& \multicolumn{4}{c}{\textbf{Video Fidelity}} \\
\cmidrule(lr){2-4} \cmidrule(lr){5-8} \cmidrule(lr){9-12}
 & UAS & IFS & VRS 
 & VQR & AES & MSM & TQ 
 & SEM & SSIM & MF & SRC \\
\midrule
\multicolumn{12}{c}{Closed-Source} \\
\midrule
Wan2.7       & \textbf{56.89} & \textbf{82.02} & 79.97 &\textbf{4.407}  &\textbf{5.077}  &0.692  &18.223  & 87.90 &0.482  &0.896  & 0.815 \\
HappyHorse1.0    & 55.18 & 76.54 & \textbf{84.52} &4.388  &5.070  &\textbf{0.710}  &\textbf{18.414}  & \textbf{92.73} &0.506  &0.886  &0.823  \\

\midrule
\multicolumn{12}{c}{Open-Source} \\
\midrule
OmniWeaving  & 30.14 & 57.18 & 61.75 & 3.660 & 4.135 & 0.709 & 15.092 & 85.05 & 0.463 & 0.891 & 0.781 \\
Kiwi         & 29.03 & 53.90 & 56.13 & 3.670 & 4.609 & 0.642 & 15.649 & 79.51 & 0.605 & 0.893 & 0.814 \\
Ditto        & 26.50 & 49.45 & 60.69 & 3.921 & 4.297 & 0.639 & 15.583 & 58.02 & 0.355 & 0.907 & 0.763 \\
Lucy         & 26.01 & 50.85 & 58.68 & 3.688 & 4.136 & 0.661 & 15.045 & 86.13 & \textbf{0.762} & 0.918 & \textbf{0.834} \\
ICVE         & 25.83 & 53.14 & 54.00 & 3.277 & 3.695 & 0.642 & 12.168 & 71.02 & 0.288 & 0.814 & 0.642 \\
ReCo         & 24.35 & 54.16 & 47.42 & 3.146 & 3.906 & 0.625 & 12.101 & 70.03 & 0.528 & 0.870 & 0.730 \\
InsV2V       & 14.61 & 37.18 & 47.36 & 3.307 & 4.327 & 0.698 & 10.501 & 77.85 & 0.280 & 0.886 & 0.740 \\
VACE         &  9.69 & 22.92 & 41.35 & 3.718 & 5.037 & 0.688 & 13.637 & 81.73 & 0.709 & \textbf{0.958} & 0.783 \\
\bottomrule
\end{tabular}
}
\caption{Quantitative results on CoVEBench across instruction compliance, video quality, and video fidelity. UAS, IFS, VRS, and SEM are checklist-based scores. All metrics are higher-is-better.}
\label{tab:main-results}
\end{table*}

\begin{enumerate}
    \item \textbf{Closed-source models tend to show stronger instruction compliance.} The two evaluated closed-source models achieve notably higher scores on checklist-based metrics, especially UAS, though this trend is based on a limited sample of proprietary models.

    \item \textbf{Current models still struggle to achieve high-quality instruction completion.}
    Although some models can follow the requested edits, their union accuracy UAS is much lower than the individual instruction-following and realism scores. This suggests that completing all required edits while maintaining physical plausibility remains challenging.

    \item \textbf{There is a clear trade-off between edit execution and content preservation.}
    Some models obtain relatively strong instruction-following performance but suffer from weaker semantic preservation. For example, Ditto achieves competitive execution-related scores, yet its SEM score is much lower, indicating that stronger edits may come at the cost of unintended changes to preserved content.

\end{enumerate}

\subsection{Further Analysis}

\paragraph{\textbf{Metric Validity Evaluation.}}

We evaluate metric validity through stability and human-alignment analyses. To assess stability, we conduct repeated checklist evaluations on OmniWeaving and Kiwi; the score variations remain strictly within 0.5 points, demonstrating high reliability. To evaluate human alignment on objective questions, we sample 100 cases selected from Wan and Kiwi. Both human evaluators and Qwen3.5-122B-A10B provide answers to all objective questions associated with these samples. The agreement rate is calculated as the proportion of questions where the model's answers exactly match the human annotations out of the total number of evaluated objective questions. Results show that Qwen3.5-122B-A10B achieves an agreement of over 93\% (Cohen's $\kappa=0.84$). At the system level, the model ranking derived from UAS perfectly matches human judgments of overall editing quality. Furthermore, pairwise comparisons (60 samples) show that all other metrics achieve over 85\% agreement with human preferences (Table~\ref{tab:metric_human_consistency}). Detailed settings are in the appendix~\ref{app:agree}.
\begin{table}[htbp]
\centering
\small
\setlength{\tabcolsep}{2pt}
\resizebox{0.6\columnwidth}{!}{
\begin{tabular}{lcccccccc}
\toprule
Metric & SEM & TQ & AES & MSM & VQR & MF & SSIM & SRC \\
\midrule
Agreement & 93.3 & 85.7 & 90.0 & 91.4 & 85.7 & 91.4 & 85.0 & 85.7 \\
\bottomrule
\end{tabular}
}
\caption{
Human preference consistency of metrics. Agreement measures whether metric-induced pairwise rankings match human preferences. 
}
\label{tab:metric_human_consistency}
\end{table}


\begin{figure*}[htbp]
    \centering
    \includegraphics[width=\linewidth]{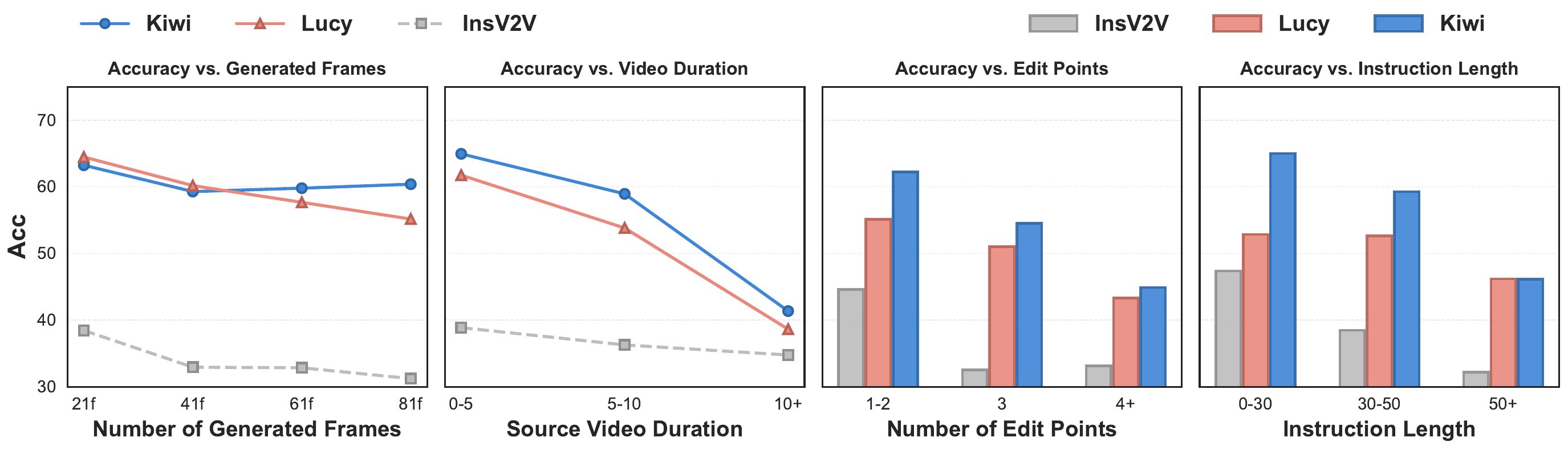}
    \vspace{-10pt}
    \caption{
    Analysis of model robustness under increasing temporal and editing complexity.
The four panels report Acc under different generated frame lengths, source video durations, numbers of edit points, and instruction lengths. Acc denotes the average accuracy across all objective checklist questions.
    }
    \vspace{-10pt}
    \label{fig:temporal_stress_test}
\end{figure*}

\paragraph{\textbf{Inference Efficiency Evaluation.}}

To assess practical deployment feasibility, we profile the inference efficiency of the evaluated models—specifically, the average time per frame and peak VRAM footprint—under a standardized configuration of $480 \times 480$ spatial resolution and 41 video frames. All evaluations are conducted on a single NVIDIA H200 GPU, with the results summarized in Table~\ref{tab:efficiency}. The comparison reveals disparities in computational requirements. For inference speed, Kiwi and Lucy demonstrate superior time efficiency, rendering them highly practical for latency-sensitive applications. Regarding memory allocation, InsV2V, ReCo, and OmniWeaving maintain optimized VRAM footprints, ensuring feasible deployment under constrained hardware. In stark contrast, models like VACE and ICVE demand excessive computational time and memory reserves, severely limiting their practical accessibility.

\begin{table}[htbp]
\centering
\vspace{-2pt}

\resizebox{0.5\columnwidth}{!}{
\begin{tabular}{lcc}
\toprule
\textbf{Method} & \textbf{Time (s/f) $\downarrow$} & \textbf{Peak VRAM (MB) $\downarrow$} \\
\midrule
Kiwi & \textbf{0.651} & 31,476.3 \\
Lucy & \underline{0.721} & 33,020.8 \\
ReCo & 2.587 & \underline{21,475.0} \\
InsV2V & 2.646 & \textbf{14,665.6} \\
OmniWeaving & 4.708 & 22,567.6 \\
ICVE & 5.621 & 68,130.6 \\
Ditto & 7.288 & 45,845.9 \\
VACE & 12.827 & 113,090.9 \\
\bottomrule
\end{tabular}

}
\vspace{-3pt}
\caption{Inference efficiency comparison. The average inference time is measured in seconds per frame (s/f) and the peak VRAM footprint is reported in MB. All models are evaluated on a single NVIDIA H200 GPU.}
\vspace{-10pt}
\label{tab:efficiency}
\end{table}

\paragraph{\textbf{Temporal Scalability Analysis.}}
We evaluate the temporal scalability of InsV2V, Kiwi, and Lucy using 300 sampled instances per model (Figure~\ref{fig:temporal_stress_test}). The analysis spans two dimensions: generated frame count, which tests consistency in long-sequence generation, and source video duration, which evaluates the capacity for handling extended input conditions. Overall, results reveal a general performance decline across both dimensions, underscoring that longer temporal spans significantly amplify the difficulty of the task. Secondary to this universal trend, we observe noticeable variance among models; most notably, Kiwi exhibits unique robustness in longer frame generation, maintaining stability while others falter.

\paragraph{\textbf{Editing Complexity Analysis.}}
To investigate the impact of editing complexity, we evaluate InsV2V, Kiwi, and Lucy across the entire dataset. The right panels of Figure~\ref{fig:temporal_stress_test} present the results categorized by the number of edit points and instruction length. Across all evaluated models, performance exhibits a clear downward trend as the number of edit points increases, indicating that complex edits remain a significant challenge. A similar degradation occurs with longer instructions, where scores drop as prompts encapsulate more intricate editing details. These findings reveal that current models are highly sensitive to both the structural and semantic complexity of the editing instructions.

\begin{figure*}[t]
    \centering
    \vspace{-10pt}
    \includegraphics[width=\linewidth]
    {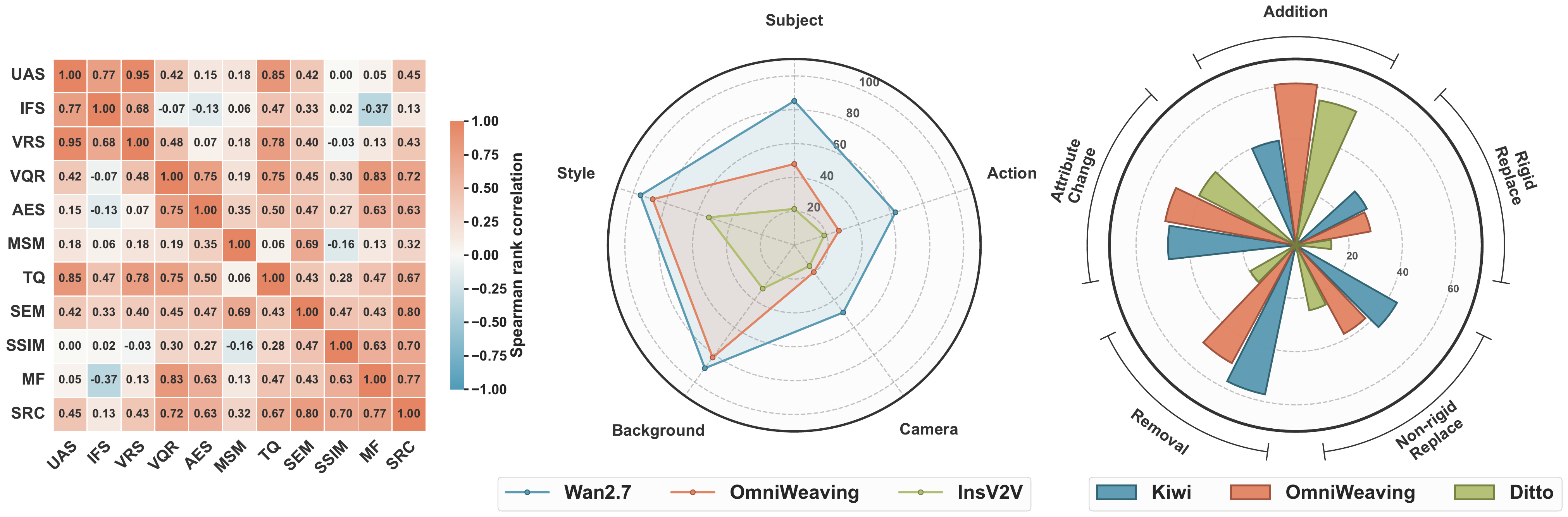}
    \vspace{-10pt}
    \caption{Metric correlation and fine-grained editing category analysis. Left: correlation among evaluation metrics across open-source models. Middle: model performance across major editing categories. Right: subject-editing breakdown across representative subject operations.}
    \label{fig:metric_category_analysis}
    \vspace{-10pt}
\end{figure*}

\paragraph{\textbf{Single-Edit Decomposition vs. Joint Editing.}}
We compare joint editing (applying all instructions simultaneously) against sequential decomposition (step-by-step editing). Joint editing consistently outperforms the sequential approach in UAS (30.63\% vs. 23.70\%), IFS (56.56\% vs. 48.68\%), and VRS (56.48\% vs. 51.62\%). The performance degradation in sequential editing stems from two primary issues: (1) Insufficient preservation: new edits frequently overwrite previous modifications; and (2) Artifact accumulation: errors in degraded intermediate videos snowball, complicating subsequent edits. Furthermore, sequential editing requires multiple inference passes, significantly increasing computational cost and latency. By processing requirements holistically, joint editing ultimately achieves superior quality and efficiency.

\paragraph{\textbf{Metric Complementarity.}}
Figure~\ref{fig:metric_category_analysis} presents the Spearman rank-correlation heatmap of the evaluation metrics. Overall, the metric suite exhibits low redundancy: strong correlations appear almost exclusively between semantically related pairs within the same dimension, while cross-dimension correlations are substantially weaker and heterogeneous. This confirms that instruction compliance, video quality, and video fidelity capture complementary aspects of model performance, and the proposed metrics collectively provide diverse, non-redundant evaluation signals.

\paragraph{\textbf{Fine-Grained Category Analysis.}}
As shown in Figure~\ref{fig:metric_category_analysis}, current models struggle with specific editing types. Notably, camera control, motion, and subject edits remain significantly harder than background or style modifications—a limitation shared even by closed-source models. Within subject edits, replacement (particularly rigid/non-rigid) proves more challenging than addition or removal. These disparities highlight how aggregate scores often mask critical flaws, underscoring the diagnostic value of our fine-grained checklist evaluation (further details in Appendix~\ref{category}).

\paragraph{\textbf{Error Analysis.}}

\begin{figure}[t]
    \centering
    \includegraphics[width=0.48\textwidth]{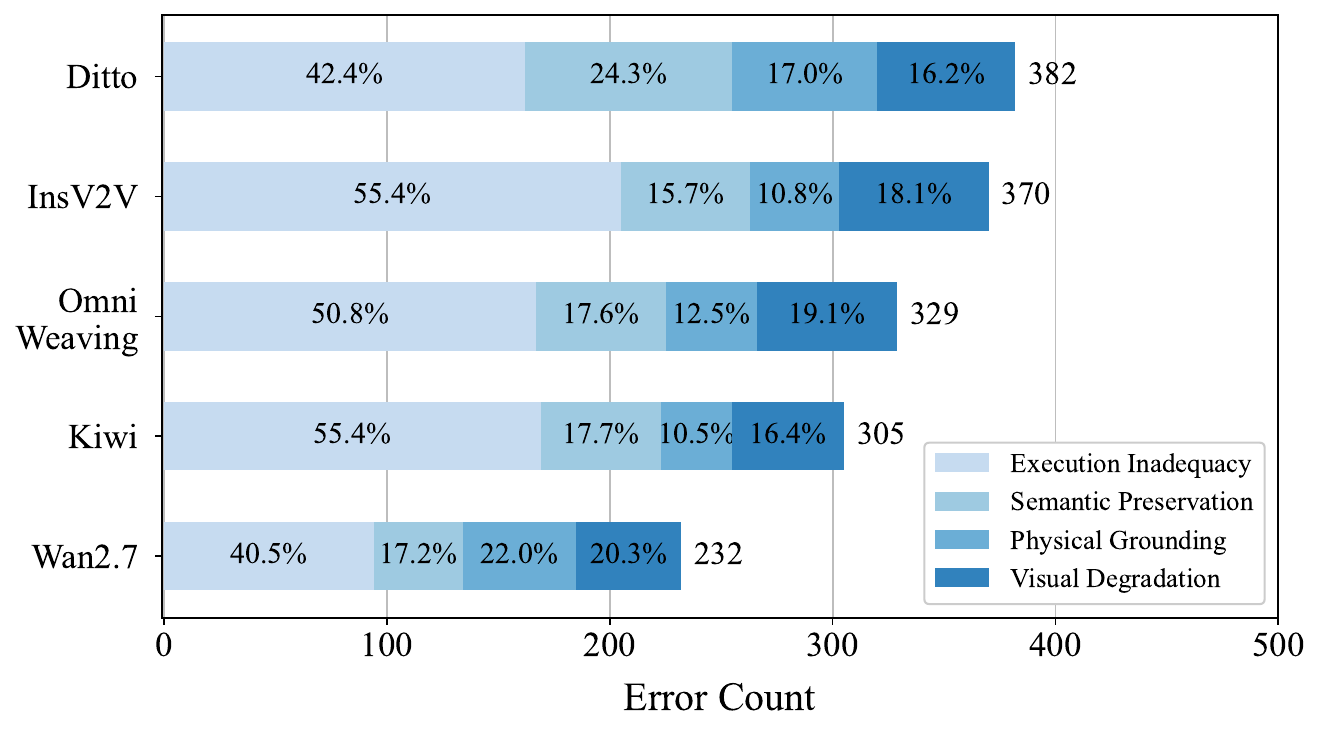}
    \vspace{-5mm}
    \caption{High-level error analysis of five video editing models. Percentages within the stacked segments show the relative proportion of each error type.}
    \vspace{-5mm}
    \label{fig:error_analysis}
\end{figure}

To systematically evaluate the limitations of current video editing approaches, we conduct a comprehensive error analysis on five recent models: Wan2.7, OmniWeaving, Kiwi, Ditto, and InsV2V. Specifically, we randomly select 100 identical samples for each model to perform a statistical evaluation. We categorize the prevalent failures into four critical types: (1) \textit{execution inadequacies} (i.e., poor instruction following or text rendering); (2) \textit{spatial entanglement}, leading to unintended modifications of non-target regions; (3) a \textit{lack of physical grounding}, resulting in unnatural motions; and (4) \textit{visual degradation}, characterized by a loss of photorealism. The statistical distribution is illustrated in Figure \ref{fig:error_analysis}. 

As indicated by the results, inadequate instruction following emerges as the most critical bottleneck that open-source models urgently need to overcome. Although closed-source models exhibit comparatively better instruction adherence, their completion rate drops significantly when handling highly complex instructions. Furthermore, we observe that editing with closed-source models frequently introduces severe physical violations and unnatural blending (e.g., "sticker-like" copy-paste artifacts), exacerbating issues related to physical grounding and visual coherence. Further qualitative examples and detailed experimental settings are provided in Appendix \ref{app:error}. Together, these persistent shortcomings severely hinder the practical application of current models, thereby motivating the development of our novel approach.

\section{Conclusion}

This paper introduces CoVEBench, a rigorous evaluation framework designed to probe the limits of compositional video editing. Utilizing a suite of automated metrics and nearly 10,000 fine-grained MLLM checklists, we provide a holistic assessment of execution accuracy, visual quality, and content preservation. Our findings expose weaknesses in current models, such as edit omissions, physical violations, and unwanted background shifts-revealing fundamental bottlenecks in their internal logic and the inability to independently process compositional spatiotemporal concepts.

\section*{Limitations}

While CoVEBench provides a comprehensive and challenging testbed for compositional video editing, it has several limitations. First, our current scope is primarily restricted to text-guided instructions. In advanced real-world workflows, creators often rely on supplementary multi-modal control signals—such as reference images, spatial bounding boxes, or audio cues—to achieve precise spatial-temporal alignment, which are not accommodated in our current framework. Second, CoVEBench serves exclusively as an evaluation benchmark; we do not propose a corresponding solution, such as an LLM-driven agent, to autonomously tackle the complex compositional tasks we identified. Finally, this work does not provide a paired large-scale training dataset, which limits the immediate ability to fine-tune and improve existing video editing models against these compositional challenges.

\bibliographystyle{plainnat}
\bibliography{references} 

\clearpage
\appendix
\section*{Appendix}

\section{The Practical Need for Compositional Video Editing}
\label{app:applications}
CoVEBench targets the practical regime in which AI video editing models must operate: edits involving \emph{multiple objects, attributes, regions, and stages} within the same clip, while everything else is faithfully preserved—what we refer to as \textbf{video editing in complex scenarios}. This capability directly underlies several real-world applications that single-edit benchmarks cannot expose:

\begin{itemize}

\item \textbf{Short-form Content and Social Media Creation}~\citep{Brooks2022InstructPix2PixLT}: Creators chain operations—replace the background, restyle the wardrobe, add a visual effect—and require that the central subject's identity, expression, and motion trajectory remain intact across the chain. Failure on the preservation side, even partially, makes the entire output unusable rather than merely imperfect.

\item \textbf{Film and TV Post-production}~\citep{Geyer2023TokenFlowCD,ku2024anyv2v}: A single shot is rarely subject to one edit. Editors routinely combine actor or wardrobe replacement, prop addition and removal, and brand-visibility control within the same clip, while preserving lighting, camera motion, and the identities of all unmodified subjects. A model that handles each operation in isolation but cannot compose them, or that drifts on what should remain unchanged, cannot enter this pipeline.

\item \textbf{Advertising Localization and Versioning}~\citep{Wang2023VideoComposerCV, ku2024anyv2v}: A single master campaign is shipped to many regional markets by swapping talent, on-screen text, product packaging, or background, while keeping camera framing, motion, and brand identity strictly constant. Each delivered version is itself a multi-region, multi-attribute edit on top of the same source video, where any unintended drift in the preserved layer breaks brand consistency across markets.

\end{itemize}

CoVEBench is therefore designed to measure whether current models can meet these compositional requirements. Every task, prompt, and metric in this appendix is constructed to stress-test model capabilities under exactly the multi-faceted editing conditions described above, rather than under the simplified single-edit protocols prevalent in prior benchmarks.

\section{Comparison with other Benchmarks}
\label{app:com}
In the main paper, we have already provided a systematic comparison with existing video editing benchmarks in terms of data scale, editing categories, instruction complexity, and evaluation protocols. Due to space limitations, we provide an additional comparison with UniVBench~\citep{wei2026univbenchunifiedevaluationvideo} in this appendix. UniVBench is a recent unified video benchmark covering six tasks, including video captioning, text-to-video generation, reference-image video generation, text-instruction video editing, reference-image video editing, and video reconstruction, aiming to evaluate broad unified video capabilities rather than specifically diagnosing complex compositional video editing. Therefore, although UniVBench includes some compositional editing instructions, its evaluation protocol is not designed from the perspective of complex editing. In particular, UniVBench relies heavily on an LLM-based agentic evaluation system that standardizes prompting, instruction parsing, and scoring across different tasks. While such a unified scorer improves scalability, it may also obscure fine-grained editing failures under coarse holistic scoring; for example, the reported performance of VACE in UniVBench does not reveal whether the model faithfully completes each individual edit, preserves unrelated content, or maintains physical plausibility. UniVBench does not explicitly decompose an editing instruction into multiple atomic edit points for separate assessment, nor does it decouple instruction following from video preservation. It also lacks dedicated judgment of physical realism and editing plausibility, such as whether generated motions obey physical constraints, whether edited objects are naturally integrated into the scene, or whether multi-edit interactions introduce artifacts. Moreover, its evaluation coverage is limited for closed-source video editing models, making it difficult to fairly assess the strongest proprietary systems under the same protocol. 

In contrast, CoVEBench is constructed specifically for complex compositional video editing, containing 416 curated source videos, 626 multi-point editing instructions, and 9,990 fine-grained checklist items, with each instruction involving approximately three atomic edit operations on average. More importantly, CoVEBench decomposes complex instructions into verifiable checklist questions and separately evaluates instruction compliance, video quality, and video fidelity. This design directly addresses the above limitations by enabling targeted diagnosis of atomic edit execution, edit interference, preservation errors, physical implausibility, and visual degradation. Therefore, while UniVBench is valuable as a broad unified video benchmark, CoVEBench provides a more targeted, fine-grained, and diagnostic evaluation framework for complex compositional video editing.

\section{Dataset Samples}
\label{app:samples}

\begin{figure*}[htbp]
\centering

\includegraphics[width=\textwidth]{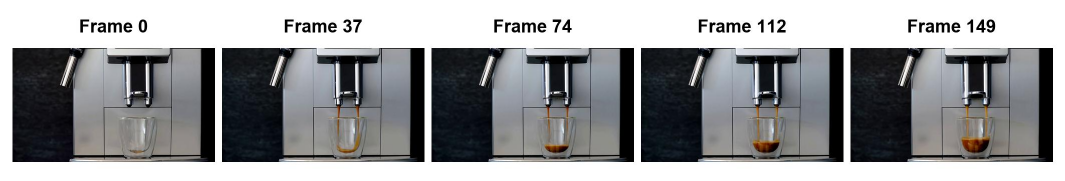}

\vspace{1mm}

\begin{tcolorbox}[mainbox, unbreakable, halign=left,
left=1mm, right=1mm, top=1mm, bottom=1mm]
\footnotesize

\textbf{Editing instruction:}
Change the black background to a white wall. Add two additional double-walled glass cups to the machine's tray, placing one on each side of the center cup to create a row of three. Fill these two new side cups with espresso and a layer of crema, while keeping the center cup positioned under the nozzles to continue receiving the pour. Ensure all other elements in the scene remain strictly unchanged.

\vspace{0.5mm}

\begin{tcolorbox}[titlebox, colback=black!60]
Addition and Placement of Side Cups
\end{tcolorbox}

\begin{enumerate}[itemsep=0.25em]

\item In Video B, are there exactly three glass cups present on the espresso machine's tray? \\
\textbf{Correct Answer:} Yes

\item In Video B, are these three cups arranged in a row? \\
\textbf{Correct Answer:} Yes

\item In Video B, are all of the cups double-layered (double-walled) glasses? \\
\textbf{Correct Answer:} Yes

\item In Video B, are any of the cups suspended in mid-air or severely blurred? \\
\textbf{Correct Answer:} No

\end{enumerate}

\begin{tcolorbox}[titlebox, colback=black!60]
Espresso Liquid and Crema
\end{tcolorbox}

\begin{enumerate}[resume, itemsep=0.25em]

\item In Video B, are the two side cups filled with dark espresso? \\
\textbf{Correct Answer:} Yes

\item In Video B, is there a visible layer of crema on the surface of the coffee in both side cups? \\
\textbf{Correct Answer:} Yes

\item In Video B, does the coffee liquid in the two side cups remain stable when no coffee is being poured into them? \\
\textbf{Correct Answer:} Yes

\item In Video B, as the espresso machine continues pouring liquid into the middle cup, is there a phenomenon where the cup is completely full but the coffee liquid does not overflow? \\
\textbf{Correct Answer:} No

\item In Video B, does the coffee liquid flowing into the middle cup appear distorted or fall unnaturally? \\
\textbf{Correct Answer:} No 

\end{enumerate}

\begin{tcolorbox}[titlebox, colback=black!60]
Background Color
\end{tcolorbox}

\begin{enumerate}[resume, itemsep=0.25em]

\item In Video B, what is the color of the background? \\
\textbf{Options:} A. White background; B. Black background. \\
\textbf{Correct Answer:} A

\end{enumerate}

\begin{tcolorbox}[titlebox, colback=black!60]
Preservation of Original Elements
\end{tcolorbox}

\begin{enumerate}[resume, itemsep=0.25em]

\item Comparing Video A and Video B, how accurately are the two streams of espresso pouring into the center cup preserved? \\
\textbf{Correct Answer:} 10

\item Comparing Video A and Video B, how well is the static camera framing and medium close-up shot preserved? \\
\textbf{Correct Answer:} 10

\item Comparing Video A and Video B, how well is the silver espresso machine's appearance and metallic texture preserved? \\
\textbf{Correct Answer:} 10

\end{enumerate}

\end{tcolorbox}

\caption{Representative dataset sample. The images display frames sampled from the original video, and the text box below presents a complete example of the corresponding evaluation checklist.}
\label{fig:dataset_sample}
\end{figure*}

To better illustrate our dataset, we present one representative example in Fig.~\ref{fig:dataset_sample}. The example contains five key frames sampled from a single video, together with a checklist of questions. These questions are grouped by specific editing points and are designed to evaluate the accuracy of the video editing---specifically, whether the target semantic changes are correctly applied while the untargeted attributes are faithfully preserved.

\section{Construction of the Test Set}
\label{app:test}

\begin{figure*}[htbp]
\centering
\includegraphics[width=\textwidth]{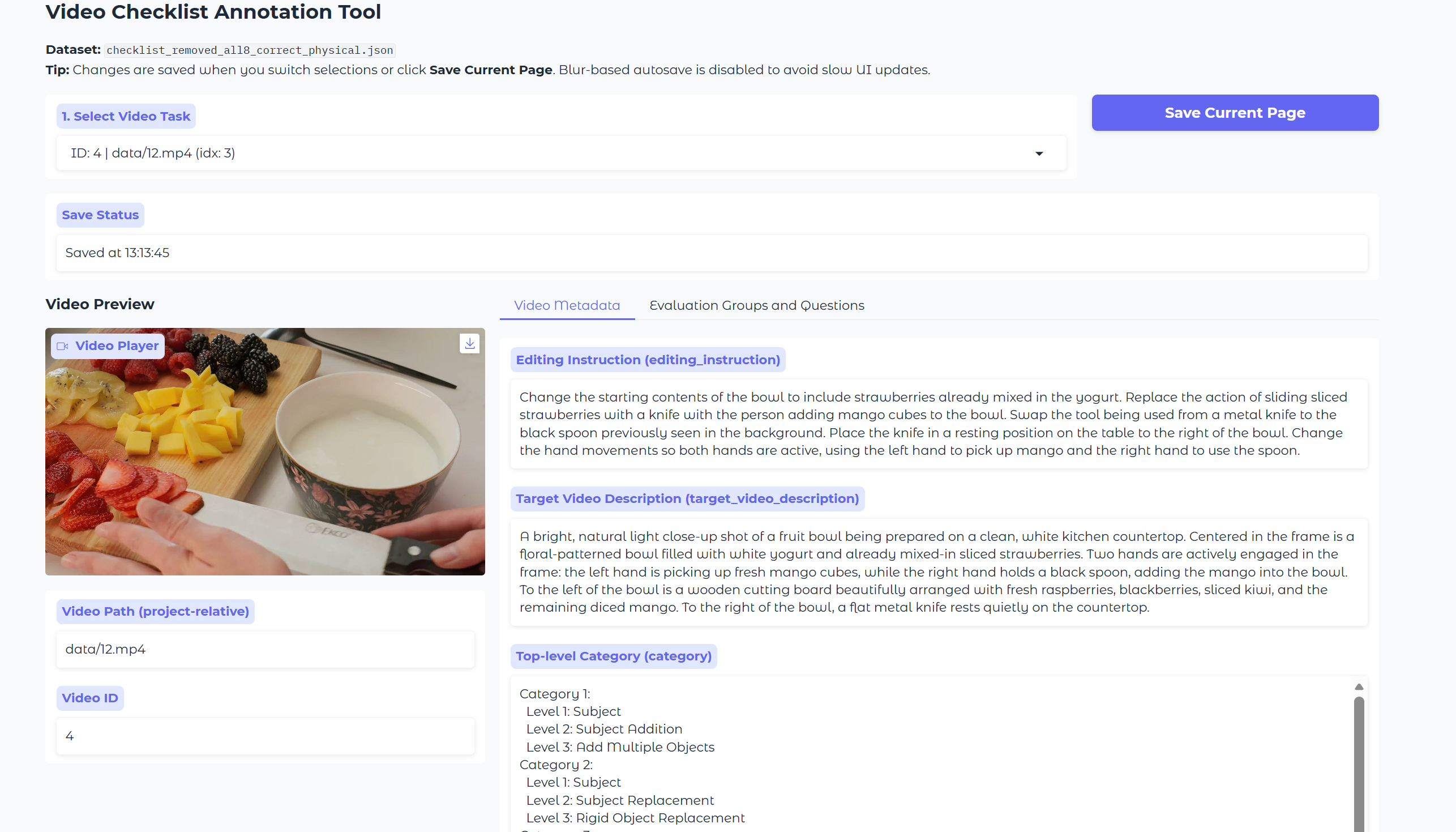}
\caption{Annotation interface used for checklist verification and refinement. Annotators review videos, editing instructions, category labels, evaluation groups, questions, options, and expected answers in a structured workflow.}
\label{fig:annotation_interface}
\end{figure*}

To ensure the quality and consistency of our benchmark, we developed a structured annotation interface for checklist verification and refinement, as shown in Fig.~\ref{fig:annotation_interface}. Annotators are presented with the source video, editing instruction, target video description, evaluation groups, question categories, question text, answer options, and expected answers in a unified interface. By design, the system inherently structures each video-editing case into multiple evaluation groups according to the intended editing operations, where each group contains fine-grained questions targeting aspects such as execution accuracy, physical logic, or semantic preservation.

The annotation process is conducted by a team of six trained annotators following a rigorous workflow. First, annotators review the editing instruction itself to verify its completeness and clarity. They then evaluate and refine the pre-generated checklist, revising the question wording, category labels, answer choices, and expected answers. Subsequently, each drafted checklist is sequentially reviewed and corrected by two annotators based on unified criteria, targeting issues such as factual errors, logical contradictions, misclassifications, excessive subjectivity, or omissions. Crucially, the review scrutinizes the semantic validity, explicitly removing items that are devoid of meaningful content or involve visual distinctions that are too subtle for humans to perceive. Any disagreements are resolved through a discussion mediated by a senior annotator. This structured workflow ensures that every question is clear, unambiguous, and aligned with the corresponding visual evidence, enabling consistent checklist construction while preserving fine-grained diagnostic information for each editing instruction.

\section{Topic Distribution of Source Videos}
\label{sec:topic-distribution}

\begin{figure}[t]
    \centering
    \includegraphics[width=0.5\linewidth]{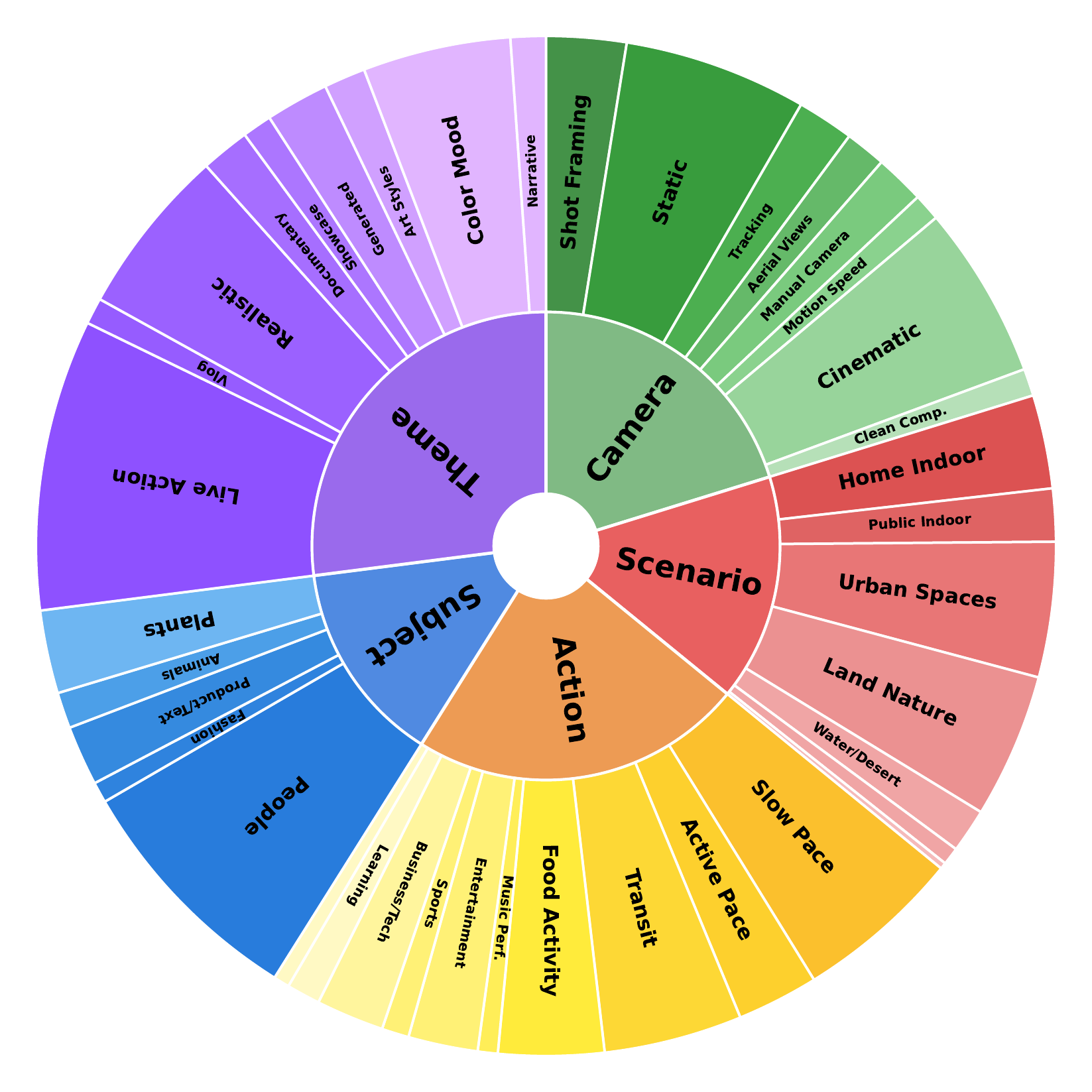}
    \caption{Topic distribution of the source videos in our dataset. The sunburst chart shows that the dataset covers diverse video domains, including scenes, subjects, actions, camera-related attributes, visual styles, and application scenarios.}
    \label{fig:video-topic-sunburst}
\end{figure}

Figure~\ref{fig:video-topic-sunburst} presents the topic distribution of our video dataset using a sunburst chart. The results show that the dataset covers a broad and diverse range of video domains, including perspective-related attributes, scene settings, actions, subjects, and thematic styles. The distribution spans real-world footage, people-centered content, indoor and outdoor environments, nature and urban scenarios, camera motion patterns, documentary and cinematic styles, as well as commercial, educational, entertainment, and creative media. This wide coverage indicates that the dataset is not concentrated in a narrow visual category, but instead provides rich variation across content types, visual styles, shooting conditions, and application scenarios, supporting robust evaluation across diverse video editing contexts.

\section{Detailed Evaluation Methodology}
\label{app:evaluation-details}

\subsection{Instruction Compliance}
We evaluate instruction compliance using three MLLM-judged metrics, assessing the edit from basic execution to absolute correctness:

\paragraph{Instruction Following (IFS)} 
IFS measures edit execution, evaluating whether the model successfully applied the fundamental edit regardless of editing quality. It is the average score of diverse checklist questions (e.g., verifying the original object is absent and the new one is present). For instance, Questions 1, 2, 3, 5, 6, and 10 in Fig.~\ref{fig:dataset_sample} serve as specific checklist items to assess these fundamental transformations.

\paragraph{Video Realism (VRS)} 
VRS assesses the visual naturalness and overall editing quality of the edited region. It averages scores from targeted questions (for instance, Questions 4, 7, 8, and 9 in Fig.~\ref{fig:dataset_sample}) detecting unnatural physical dynamics or conspicuous AI-generated artifacts caused by the edit. \textbf{Benefit:} It goes beyond mere execution to evaluate whether the modification is natural, coherent, and artifact-free.

\paragraph{Union Accuracy (UAS)} 
UAS is a stringent, union metric evaluating a model's ability to completely execute a specific edit point. An edit receives a score of 1 \textit{only} if both instruction and realism queries are correct (i.e., it perfectly answers all associated IFS and VRS questions); otherwise, it scores 0. For instance, in the ``Addition and Placement of Side Cups'' edit shown in Fig.~\ref{fig:dataset_sample}, the model earns a score of 1 if and only if it correctly answers all four corresponding questions. \textbf{Benefit:} As the strictest metric, UAS guarantees that a high score reflects a flawless edit—both accurately executed and qualitatively perfect.

\subsection{Video Quality}

\paragraph{Comprehensive Quality (VQR)}
VQR evaluates the overall perceptual quality of the edited video. We apply VisualQuality-R1 to 10 sampled frames from each edited video. VisualQuality-R1 is a reasoning-based no-reference image quality assessment model that estimates human-aligned visual quality without requiring reference images. The VQR score is computed by averaging the VisualQuality-R1 scores over the 10 sampled frames. \textbf{Benefit:} As the main indicator of the Video Quality dimension, VQR reflects the overall clarity, naturalness, and perceived quality of the edited video as judged by human viewers.

\paragraph{Aesthetics (AES)}
AES evaluates the frame- and keyframe-level visual appeal of the edited video using Aesthetic Predictor v2.5. For each video, we score both the keyframes and 10 uniformly sampled frames, then compute AES as the average of all keyframe and sampled-frame aesthetic scores. \textbf{Benefit:} AES captures whether the edited result is visually pleasing in terms of composition, color, lighting, and overall photographic quality, rather than only measuring task completion.

\paragraph{Motion Smoothness (MSM)}
MSM measures the temporal motion smoothness of the edited video itself. We uniformly sample $T=10$ frames from each edited video and estimate dense optical flow between adjacent frames. Let
\[
F_t^{edit}=\mathrm{Flow}(I_t^{edit}, I_{t+1}^{edit})
\]
denote the optical-flow field between two consecutive sampled frames. We measure motion jitter by comparing consecutive flow fields:
\[
J_t(p)=
\frac{
\|F_{t+1}^{edit}(p)-F_t^{edit}(p)\|_2
}{
\|F_{t+1}^{edit}(p)\|_2+\|F_t^{edit}(p)\|_2+\epsilon
}.
\]
The video-level motion smoothness score is computed as
\[
\mathrm{MSM}
=
1-
\frac{1}{T-2}
\sum_{t=1}^{T-2}
\frac{1}{|M_t|}
\sum_{p\in M_t}
\min(1,J_t(p)),
\]
where $M_t$ denotes pixels with non-negligible motion. A higher MSM indicates smoother and more temporally stable motion in the edited video.

\paragraph{Technical Quality (TQ)}

TQ measures the perceived technical distortions of the edited video using the technical score from DOVER++. DOVER++ is a no-reference video quality assessment model whose technical branch focuses on distortions such as blur, noise, compression artifacts, and temporal flickering. \textbf{Benefit:} TQ evaluates whether the edited video remains technically clean and watchable after editing, regardless of whether the edit is semantically correct.


\begin{table*}[t]
\centering

\small
\resizebox{\textwidth}{!}{
\begin{tabular}{lcccccccc}
\toprule
Metric & Kiwi & Lucy & ICVE & InsV2V & VACE & ReCo & Ditto & OmniWeaving \\
\midrule
Text--Video CLIP $\uparrow$
& 22.7068 & 23.3102 & 22.4380 & 23.0057 & 22.5734 & 22.1487 & 22.6583 & 23.2970 \\
\bottomrule
\end{tabular}
\vspace{-20pt}
}
\caption{Text--Video CLIP Fails to Distinguish Fine-Grained Instruction Compliance.}
\vspace{-10pt}
\label{tab:clip_instruction_compliance}
\end{table*}

\subsection{Video Fidelity}

\paragraph{Semantic Consistency (SEM)}

SEM evaluates whether the content that should remain unchanged is preserved after editing. For each editing case, we construct preservation-oriented checklist questions for the main unedited elements, such as Q11--Q13 in Fig.~\ref{fig:dataset_sample}, which verify the consistency of non-target content, including objects, backgrounds, attributes, actions, camera motion, spatial relationships, and other instruction-irrelevant semantics. We provide the MLLM with the source video, the edited video, the editing instruction, and each checklist question. The source and edited videos enable direct before-and-after comparison, while the editing instruction specifies which changes are intended and which content should be preserved. This instruction-aware setup is important for edits such as style transfer, where large visual changes are expected and should not be penalized as semantic inconsistency. The MLLM scores each preservation-oriented question, and the SEM score is computed as the average score over all such questions. \textbf{Benefit:} SEM measures both static semantic preservation, such as objects, attributes, backgrounds, and spatial layouts, and dynamic semantic preservation, such as actions, camera motion, and temporal relationships.

\paragraph{Structural Fidelity (SSIM)}
Structural Fidelity measures frame-level layout and structure preservation between the source video and the edited video. Following the VEditBench structural similarity setting, we uniformly sample $T=10$ corresponding frames from both videos, convert them to grayscale, and compute the Structural Similarity Index (SSIM) on each frame pair:
\[
\mathrm{SSIM}(x,y)=
\frac{(2\mu_x\mu_y+C_1)(2\sigma_{xy}+C_2)}
{(\mu_x^2+\mu_y^2+C_1)(\sigma_x^2+\sigma_y^2+C_2)} .
\]
The video-level score is obtained by averaging over all sampled frame pairs:
\[
\mathrm{SF}=\frac{1}{T}\sum_{t=1}^{T}
\mathrm{SSIM}\left(I_t^{src}, I_t^{edit}\right).
\]
\textbf{Benefit:} This metric captures whether the global layout, shapes, and spatial structure of the original video are preserved after editing. However, since SSIM is highly sensitive to background alterations, we exclude cases involving background changes when calculating this metric to ensure a fair evaluation of structure preservation.

\paragraph{Motion Fidelity (MF)}
MF measures trajectory-level motion preservation using CoTracker. Given sampled frames from the source and edited videos, CoTracker extracts point trajectories. For each trajectory, we summarize its average position $p_i$ and average velocity $v_i$. The matching cost between source trajectory $i$ and edited trajectory $j$ is
\begin{equation*}
\begin{aligned}
c_{ij} &= \alpha \frac{\|p_i^{src}-p_j^{edit}\|_2}{d_{\max}} \\
&\quad + (1-\alpha) \left( 1- \frac{v_i^{src}\cdot v_j^{edit}}{\|v_i^{src}\|_2\|v_j^{edit}\|_2+\epsilon} \right),
\end{aligned}
\end{equation*}
where $d_{\max}$ is the image diagonal and $\alpha=0.5$. We use Hungarian matching to find the best trajectory correspondence $\pi^*$, and define
\[
\mathrm{MF}
=
1-
\frac{1}{N}
\sum_{i=1}^{N}
c_{i,\pi^*(i)} .
\]
A higher MF indicates that the edited video better preserves the source-video motion trajectories.

\paragraph{Static Region Consistency (SRC)} 

SRC performs mask-based feature comparison for unchanged areas to evaluate whether preserved entities maintain their visual identity after editing. We obtain masks for preserved static entities using Grounding DINO and SAM2, crop the corresponding masked regions, and compute DINOv2 feature cosine similarity between the source and edited videos. Crucially, we introduce a filtering mechanism for this metric. Since mask-based similarity relies on spatial and textural alignment, it is ill-suited for samples involving significant camera movements (e.g., zooming in/out), global stylistic transformations, or excessive visual alterations. Such samples are explicitly excluded from the SRC calculation to prevent skewed results. \textbf{Benefit:} By filtering out these incompatible cases, SRC provides a highly reliable measure of subject-level preservation, accurately reflecting whether objects or characters that should remain unchanged retain their original appearance despite surrounding edits.

\begin{table*}[t]
\centering
\small
\renewcommand{\arraystretch}{1.15}

\resizebox{\textwidth}{!}{
\begin{tabular}{lccccc}
\toprule
\textbf{Model}
& \textbf{Resolution}
& \textbf{Min Frames}
& \textbf{Max Frames}
& \textbf{Default / Used Frames}
& \textbf{Frame Constraint} \\
\midrule

OmniWeaving
& $848 \times 480$
& 33
& 161
& Original
& $4k+1$ \\

Kiwi
& $\leq 720$p
& 33
& 161
& Original
& $4k+1$ \\

ReCo
& $\leq 480 \times 832$
& 17
& 257
& Original
& $4k+1$ \\

VACE
& 720p
& 33
& 81
& Original
& $4k+1$ \\

Ditto
& $480 \times 832$
& 33
& 257
& Original
& $4k+1$ \\

ICVE
& $\leq 480 \times 832$
& 17
& 209
& Original
& $4k+1$ \\

Lucy
& $\leq 1280 \times 720$
& 17
& 257
& Original
& $4k+1$ \\

InsV2V
& $384 \times 384$
& 0
& 64
& 64
& -- \\

\bottomrule
\end{tabular}
}
\caption{Inference settings for the evaluated video editing models. ``Resolution'' denotes the output resolution or the maximum resolution allowed by each model's official pipeline. ``Min Frames'' and ``Max Frames'' indicate the valid input frame range after preprocessing. ``Default / Used Frames'' specifies whether the model follows the original input length or uses a fixed frame count. ``Frame Constraint'' reports additional requirements imposed by the model, such as the $4k+1$ frame-count rule.}
\label{tab:model_inference_settings}
\end{table*}

\subsection{Discussion on metrics}

\paragraph{\textbf{Limitations of Text--Video CLIP for Instruction Compliance.}}
As shown in Table~\ref{tab:clip_instruction_compliance}, relying on Text--Video CLIP for evaluation exhibits four major limitations: 
(1) \textbf{Marginal variance:} The scores fluctuate within a surprisingly narrow range (22.15 to 23.31), making it difficult to distinguish true performance gaps. 
(2) \textbf{Inconsistency with human judgment:} The metric ranking contradicts subjective evaluations. For example, VACE outscores ReCo (22.57 vs. 22.15) despite VACE largely failing to execute edits in practice. 
(3) \textbf{Entanglement of compliance and preservation:} By measuring the similarity between the target prompt and the edited video, CLIP fails to decouple actual instruction compliance from source video preservation. High scores often originate from the unaltered content (e.g., original background and objects) rather than the successful execution of the edit. This perfectly explains why conservative models like VACE still receive competitive scores. 
(4) \textbf{Coarse granularity:} It only yields a single global score without diagnosing which specific categories of edits (e.g., color, action, or object) failed. 
These flaws suggest that global text--video alignment is inadequate for fine-grained video editing evaluation, whereas our explicitly decoupled, checklist-based metric provides discriminative and interpretable results that well align with human perception.

\paragraph{Rationale for the Evaluation Metrics}

Existing evaluation paradigms relying on global similarity inherently penalize intentional, instruction-driven modifications. This flaw is particularly severe in complex video editing, where visual and structural transformations are drastic. To address this, we reconstruct the evaluation paradigm by explicitly decoupling intended changes from expected preservation. At the semantic level, we propose an instruction-aware checklist (SEM) that parses the prompt to selectively verify only the unedited entities and backgrounds, avoiding blind whole-frame comparisons. Furthermore, for low-level dynamic metrics like Motion Fidelity (MF) and Static Region Consistency (SRC), we introduce rigorous sample filtering. By excluding test cases involving instruction-dictated motion changes or massive global stylizations, we confine similarity calculations strictly to contexts where visual preservation is logically expected. This targeted, "calculate-as-needed" design effectively mitigates the inherent biases of traditional metrics, ensuring a highly reliable assessment.

\section{Experiment}

\subsection{Experiment Settings.}
We evaluate eight representative open-source video editing models, including InsV2V~\citep{ICLR2024_48a13e12}, VACE~\citep{Jiang_2025_ICCV}, Lucy ~\citep{decart2025lucyedit}, ICVE~\citep{liao2025context}, Ditto~\citep{bai2025scalinginstructionbasedvideoediting}, ReCo~\citep{Zhang2025RegionConstraintIG}, OmniWeaving~\citep{pan2026omniweavingunifiedvideogeneration}, and Kiwi~\citep{kiwiedit}. All experiments are performed on a single NVIDIA H200 GPU. For each model, we follow its official inference pipeline and use the publicly released checkpoints and recommended default hyperparameters whenever available. Each model's output resolution follows its officially recommended setting, and the number of generated frames is matched to the input sequence whenever permitted by the model's frame constraints. When a model imposes a specific frame-count range or mathematical requirement, we preprocess the input videos accordingly, such as sampling frames to satisfy the required format. The detailed parameter configurations for all evaluated models are summarized in Table~\ref{tab:model_inference_settings}.

\label{category}
\begin{table*}[t]
\centering

\resizebox{\linewidth}{!}{
\begin{tabular}{lcccccccccc}
\toprule
\textbf{Model} &
\textbf{Subj. Add} &
\textbf{Subj. Rem.} &
\textbf{Subj. Rep.} &
\textbf{Subj. Attr.} &
\textbf{Bg.} &
\textbf{Camera} &
\textbf{Motion} &
\textbf{Position} &
\textbf{VFX} &
\textbf{Style} \\
\midrule
Ditto        & 55.11 & 19.66 & 18.52 & 40.85 & 67.07 & 19.32 & 25.62 & 33.33 & 66.42 & 82.58 \\
ICVE         & 38.64 & 58.97 & 51.39 & 48.28 & 51.63 & 23.05 & 31.88 & 29.13 & 41.79 & 32.58 \\
InsV2V       & 21.02 & 18.80 & 19.44 & 23.61 & 31.71 & 15.25 & 18.60 & 22.82 & 31.34 & 53.03 \\
Kiwi         & 39.77 & 57.26 & 36.11 & 48.01 & 72.76 & 20.00 & 25.62 & 27.63 & 46.27 & 78.03 \\
Lucy         & 41.48 & 23.08 & 39.81 & 47.48 & 60.16 & 16.61 & 25.05 & 33.93 & 55.22 & 55.30 \\
OmniWeaving  & 60.80 & 50.43 & 32.41 & 49.60 & 79.67 & 19.66 & 27.70 & 38.14 & 51.49 & 87.88 \\
ReCo         & 50.00 & 55.56 & 51.85 & 42.44 & 43.50 & 16.27 & 31.50 & 38.14 & 47.01 & 80.30 \\
VACE         & 10.80 &  8.55 &  5.56 & 11.94 & 21.95 & 13.22 & 12.52 & 12.91 & 32.09 & 32.58 \\
\bottomrule
\end{tabular}
}
\caption{Category-level union accuracy (UAS) of open-source video editing models. Subj. Add/Rem./Rep./Attr.'' denote subject addition, removal, replacement, and attribute modification, respectively. Bg.'' stands for background editing, and ``VFX'' for visual effects.}
\label{tab:opensource_category_uacc}
\end{table*}

\subsection{Agreement Experiment Settings.}
\label{app:agree}

We conduct a human preference alignment study to evaluate whether automatic metrics produce pairwise preferences that are consistent with human judgments. The study uses 60 pairwise comparisons among four models: VACE, OmniWeaving, Kiwi, and Ditto. For each comparison, annotators are shown the source video, the editing instruction, and two anonymized edited outputs (labeled as Video A and Video B) side by side. Crucially, the annotators are completely blind to the automatic metric scores. They are asked to provide their visual preference by choosing from three options: Video A is better in this metric, Video B is better in this metric, or Tie (Unclear). We explicitly include the "Tie" option because, in many cases, the visual differences or specific quality flaws between the two outputs are too subtle for the human eye to definitively distinguish a clear winner.

For each metric, the output with the higher score is regarded as the metric-preferred result. We then calculate the alignment between human preferences and metric preferences. If the human's clear preference (A or B) matches the metric's preference, it is counted as an agreement ($N_{\text{correct}}$); if they contradict, it is counted as a disagreement ($N_{\text{incorrect}}$). For cases where human annotators select Tie ($N_{\text{unclear}}$), we count the case as an agreement ($N_{\text{unclear-as-tie}}$) only when the absolute score difference between the two outputs is sufficiently small ($|s_A - s_B| \le \tau$). This indicates that the metric lacks a strong preference and effectively treats the pair as a tie, mirroring human visual judgment. Conversely, if the difference is large ($|s_A - s_B| > \tau$), it means the metric explicitly favors one side while humans find them visually indistinguishable, which is not counted as an agreement.

To handle the vastly different scales of various metrics (e.g., CLIP, SSIM, and AES), the tie threshold $\tau$ is not a manually fixed value. Instead, it is adaptively determined for each metric based on the overall proportion of Tie (Unclear) judgments. Specifically, if $X\%$ of the total human judgments are labeled as Tie, we define $\tau$ as the $X$-th percentile of all pairwise absolute score differences for that specific metric.
Formally, for each metric, the alignment agreement is computed as:
$$\text{Agreement} =
\frac{
N_{\text{correct}} + N_{\text{unclear-as-tie}}
}{
N_{\text{correct}} + N_{\text{incorrect}} + N_{\text{unclear}}
}.$$

\subsection{Category Analysis.}

\begin{figure}[htbp]
    \vspace{-10pt} 
    \centering
    \includegraphics[width=0.5\linewidth]{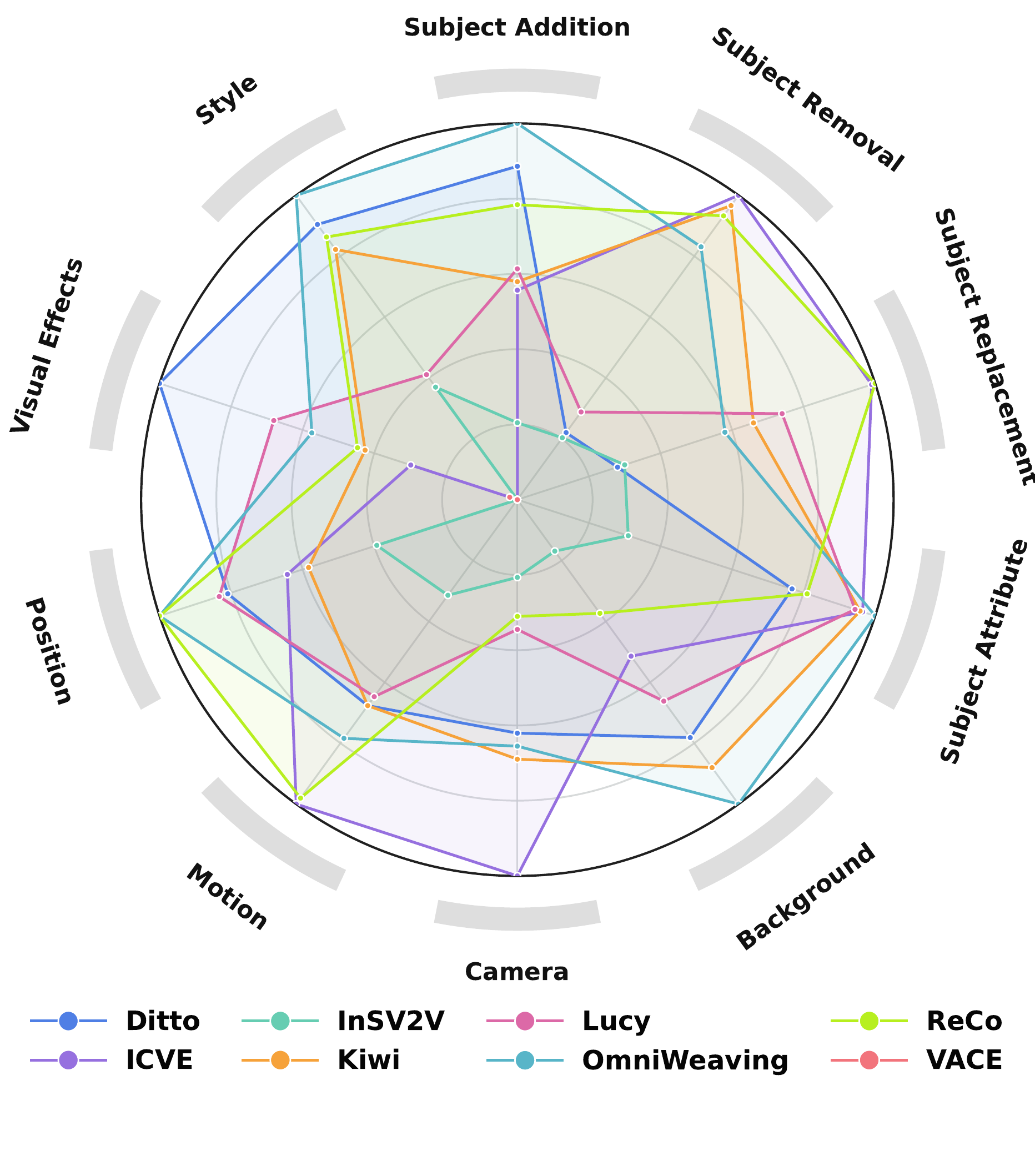}
    \vspace{-10pt} 
    \caption{Relative category-level capabilities of open-source video editing models. Axes are min-max normalized by UAS to highlight relative inter-model differences rather than absolute accuracy.}
    \label{fig:category_uacc_radar}
    \vspace{-12pt} 
\end{figure}

Table~\ref{tab:opensource_category_uacc} reports the category-level union accuracy of open-source video editing models. Overall, the results reveal substantial performance variation across editing categories. Models generally achieve higher accuracy on style transfer, background editing, and visual effects, while camera control, motion editing, and fine-grained position editing remain much more challenging. For subject-related edits, performance also varies sharply across operation types: subject addition and attribute modification are relatively easier for several models, whereas subject removal and replacement expose larger model-specific weaknesses. OmniWeaving obtains the strongest results on subject addition, background editing, position editing, and style transfer, while ReCo and ICVE perform competitively on subject replacement and removal. In contrast, VACE consistently underperforms across nearly all categories. These results suggest that current open-source models are still far from uniformly reliable: they may succeed on visually global transformations such as style or background changes, but struggle with precise spatiotemporal control, object-level manipulation, and compositional instructions involving multiple localized edits.

As shown in Figure~\ref{fig:category_uacc_radar}, which normalizes scores within each axis to highlight relative capability, open-source models exhibit highly distinct capability profiles. Rather than a single model dominating universally, the visualization reveals stark comparative advantages: models like OmniWeaving and Ditto establish the upper bounds in background and style edits, whereas ICVE and ReCo push the frontiers in challenging object-level manipulations like subject removal and replacement. This complementary nature underscores that current models are highly specialized.

\subsection{Error Analysis.}
\label{app:error}

\begin{figure*}[t]
    \centering
    \includegraphics[width=\linewidth]{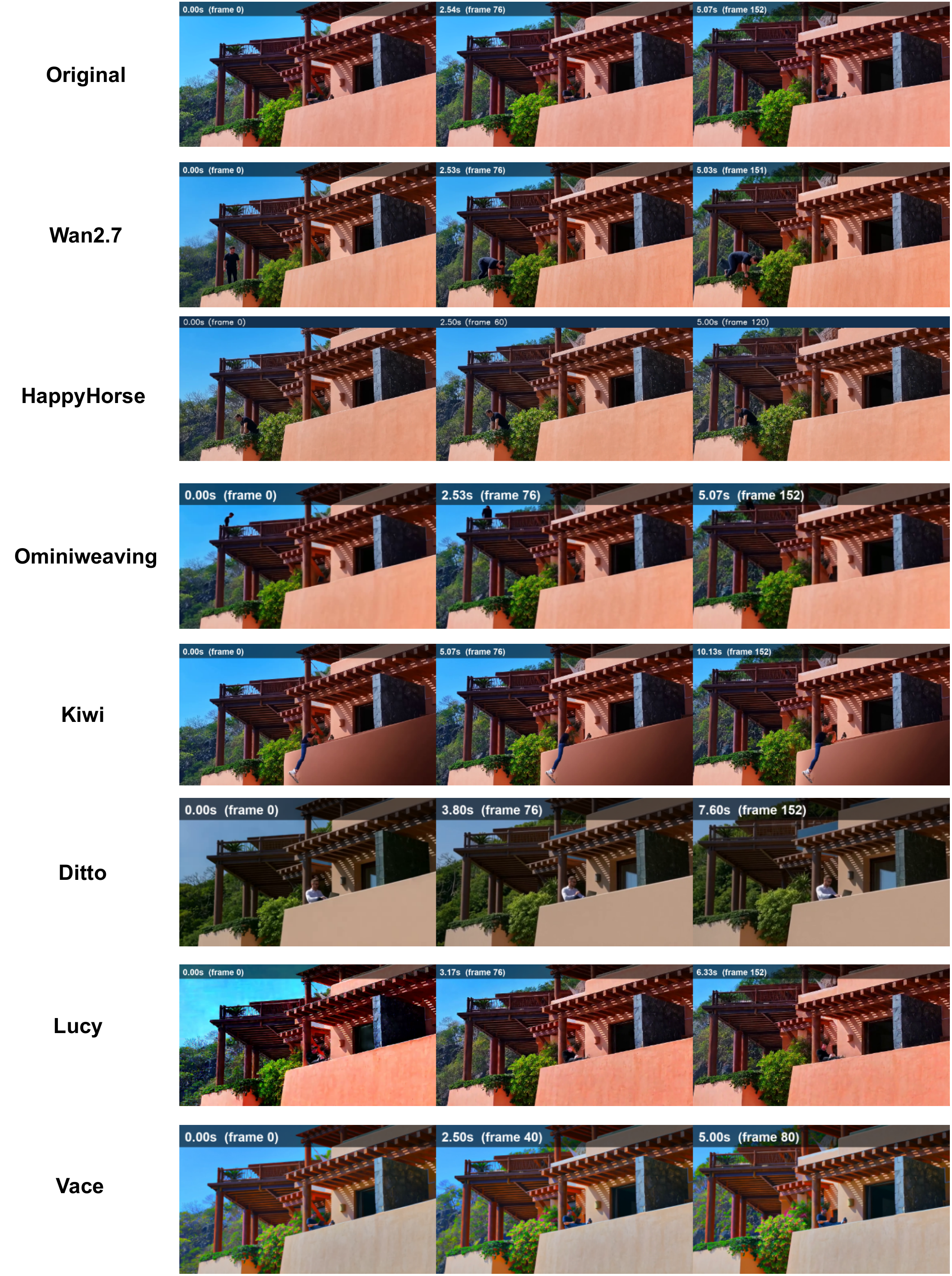}
    \caption{Qualitative comparison of different video editing models on a representative balcony-editing case. Each row shows three sampled frames from one model output, corresponding to the beginning, middle, and end of the generated video.}
    \label{fig:qualitative_error_analysis}
\end{figure*}

\subsubsection{Qualitative error analysis.}

Figure~\ref{fig:qualitative_error_analysis} shows a representative balcony-editing case, where the instruction requires changing the man's action from sitting to standing up and leaning against the pillar, shifting his relative position to the far-left side of the balcony, removing the laptop computer from the scene, making the camera static, and maintaining all other elements unchanged. Among the compared models, HappyHorse 1.0 fails to realize the actions of standing up and leaning against the pillar; it only accomplishes the downward-looking posture and the positional shift, and the camera does not remain static. Wan2.7 also fails to keep the camera fully static, cannot clearly transform the subject into a standing and leaning posture, and introduces unnatural motion ghosting around the person. In contrast, the open-source models struggle to handle such a complex multi-constraint instruction. Only Ominiweaving and Reco attempt to move the subject leftward, but both place the person toward the upper-left region and substantially alter the subject appearance and surrounding content. Kiwi introduces severe AI artifacts, making the edited person barely recognizable as a human. The remaining models show little effective editing ability and mainly blur or degrade the original video. Even the basic requirement of removing the laptop is difficult for many models to satisfy. This case suggests that complex video editing, especially instructions involving multiple challenging edit types, remains a major challenge for current models.

\subsubsection{Quantitative error analysis.} 

The counting criteria for the errors discussed in the main text are defined as follows:
\begin{itemize}
    \item \textbf{Execution Inadequacies:} This metric captures poor instruction following or incomplete edits. We measure this by verifying the completion status of specific editing points. For each editing point, we examine all related Instruction Following Score (IFS) questions. If even a single IFS-related question is marked as uncompleted, we consider the instruction following for that editing point to be problematic, and the error count increases by 1.
    \item \textbf{Spatial Entanglement:} This occurs when editing leads to unintended modifications in non-target regions. We quantify this by evaluating preservation-related questions. If the preservation score for a given region is less than or equal to 5, we determine that a severe spatial entanglement issue has occurred, adding 1 to the error count.
    \item \textbf{Lack of Physical Grounding \& Visual Degradation:} To assess unnatural motions (physical grounding) and the loss of photorealism (visual degradation), we leverage the advanced reasoning capabilities of the Qwen3.5-122B-A10B. We provide the model with the input prompts and the edited videos to automatically diagnose and list existing flaws. If the model identifies unnatural physical dynamics or visual artifacts/incoherence in a video, the respective error count increases by 1.
\end{itemize}

\section{Category taxonomy}
\label{app:category}

We design a category framework around mainstream video editing operations, covering the core edit types commonly used in controllable video generation: subject manipulation, background modification, style transfer, motion control, spatial repositioning, camera editing, and visual effects (see Table \ref{tab:category_framework}). To comprehensively address a wide range of highly challenging, real-world editing scenarios, this taxonomy is deeply anchored in practical editing intents. By assigning each evaluation question to the category that best reflects its primary semantic target, our framework enables a fine-grained, diagnostic analysis of model capabilities across diverse editing dimensions.

\clearpage

\begin{table*}[t]
\centering
\small
\renewcommand{\arraystretch}{1.18}

\begin{tabular}{>{\raggedright\bfseries}p{0.13\textwidth}
                >{\raggedright\arraybackslash}p{0.23\textwidth}
                >{\raggedright\arraybackslash}p{0.58\textwidth}}
\toprule
Level 1 & Level 2 & Level 3 / Description \\
\midrule
\multirow{4}{*}{Subject}
& Subject Addition & Adding one or multiple subjects or objects into the video. \\
& Subject Removal & Removing one or multiple subjects or objects from the video. \\
& Subject Replacement & Replacing the original subject with a rigid or non-rigid object, character, animal, or other entity. \\
& Subject Attribute Modification & Editing subject attributes, including age, color, expression, hairstyle, material, outfit, OCR/text, posture, size, etc. \\
\midrule
\multirow{3}{*}{Background}
& Scene Change & Replacing or modifying the scene, such as indoor-to-outdoor changes, seasonal changes, background replacement, or environmental setting changes. \\
& Lighting Change & Modifying illumination, color tone, time of day, brightness, cinematic lighting, sunset, nighttime, warm/cold tone, or related lighting conditions. \\
& Weather Change & Changing weather or atmospheric conditions, including sunny, rainy, snowy, foggy, windy, overcast, dusty, or thunderstorm scenes. \\
\midrule
Style
& Art Style Transfer & Transforming the visual style, such as oil painting, watercolor, sketch, cyberpunk, Studio Ghibli, American comic, retro, pixel art, claymation, black-and-white, CG rendering, or 3D cartoon style. \\
\midrule
\multirow{3}{*}{Motion}
& Motion Type & Editing the action or movement pattern, such as walking, running, jumping, waving, drinking, reading, flying, pouring, typing, spinning, or other concrete actions. \\
& Motion Direction & Modifying the direction of movement, such as forward, backward, leftward, rightward, upward, downward, or across-frame motion. \\
& Motion Speed and Amplitude & Editing motion speed, intensity, temporal order, or magnitude, including slow motion, speed-up, reverse motion, reduced motion, increased intensity, or larger/smaller movement amplitude. \\
\midrule
\multirow{2}{*}{Position}
& Relative Position & Changing spatial relationships between subjects, objects, and background elements, such as placing an object beside, above, behind, in front of, or closer to another element. \\
& Frame Layout & Changing the on-screen placement or composition, such as center, left side, right side, foreground, background, corner placement, frame alignment, or screen-position swapping. \\
\midrule
\multirow{4}{*}{Camera}
& Camera Movement & Editing camera motion, including pan, tilt, zoom in, zoom out, tracking, follow, static camera, or related camera movement changes. \\
& Camera Angle & Editing viewpoint or viewing angle, such as low angle, high angle, front view, side view, back view, or eye-level view. \\
& Shot Scale & Changing shot scale or framing distance, including close-up, medium shot, wide shot, or long shot. \\
& Lens \& Focus & Editing optical properties, especially shallow or deep depth of field and focus-related changes. \\
\midrule
\multirow{2}{*}{Visual Effects}
& Action VFX & Adding effects directly associated with subject actions, such as glowing trails, sparks, impact effects, magical aura, motion blur, splash, smoke, or energy effects. \\
& Environmental VFX & Adding scene-level effects, such as falling snow, floating particles, embers, fireflies, lens flare, heat haze, mist, lightning, digital glitch, water ripples, or environmental glow. \\
\bottomrule
\end{tabular}
\caption{Category taxonomy used in our benchmark.}
\label{tab:category_framework}
\end{table*}

\onecolumn
\section{Prompts}
\label{sec:prompts}

\subsection{Compositional Editing Instruction Generation}
\label{app:instruction_gen}

To produce compositional editing instructions, we first manually formulate 83 distinct category combinations drawn from the seven CoVEBench editing dimensions. Each combination specifies 2--4 fine-grained editing operations at the Level-2 taxonomy granularity (e.g., \textit{Subject Attribute Modification + Scene Change + Camera Movement}, or \textit{Subject Replacement + Motion Type + Action VFX}), ensuring that the resulting instructions reflect realistic and diverse multi-point editing workflows.

At prompt construction time, a preprocessing script dynamically samples 5 combinations from the pool of 83 for each source video, along with 5 corresponding few-shot input--output examples (one per sampled combination). These 5 combinations and their examples are then injected into the system prompt before it is sent to the MLLM. This dynamic rotation ensures that no single model call is biased toward a fixed subset of editing patterns, and that the resulting instruction distribution covers the full breadth of the 83-combination pool across the dataset.

The MLLM is then prompted to select the one combination most suitable for the given source video and synthesize a single cohesive multi-point editing instruction accordingly. After generation, all outputs undergo a manual review to remove inappropriate or repetitive results.

An illustrative example of the dynamically assembled prompt is shown below. The placeholders \texttt{[Combination \$i]} and \texttt{[Few-shot Example \$i]} denote the 5 combinations and their corresponding examples that are injected at runtime by the preprocessing script.

\begin{lstlisting}[style=promptstyle]
SYSTEM PROMPT: Compositional Editing Instruction Synthesizer

ROLE
You are a specialized AI that converts video editing requests into structured JSON instructions. Your goal is to analyze a source video scene and produce a clear, cohesive, multi-point edit that combines operations drawn from the seven CoVEBench editing dimensions. The resulting instruction must encode MULTIPLE atomic edits (typically 2-5, averaging around 4) that remain visually and logically coherent.

CORE DIRECTIVES

1) JSON OUTPUT ONLY. Return a single valid JSON object. No extra text, no markdown formatting outside the JSON block.

2) SCENE DESCRIPTION. The "original_description" must clearly describe the source video: main subjects (people / objects / animals), their specific actions and facial expressions, the environment and lighting, relative spatial layout, and camera framing.

3) COMBINATION SELECTION (Mandatory). You will be provided with 5 candidate combinations below, each specifying 2-4 fine-grained editing operations. Select the ONE combination that best suits the source video scene, and instantiate the corresponding atomic edits into a single cohesive instruction. On average the final instruction should specify around 4 atomic edits in total.

CANDIDATE COMBINATIONS
- Combination 1: [Combination $1]
- Combination 2: [Combination $2]
- Combination 3: [Combination $3]
- Combination 4: [Combination $4]
- Combination 5: [Combination $5]

APPROVED CATEGORIES (CoVEBench 7-D Taxonomy)

A. Subject

   - Addition / removal of a subject.
   - Replacement (e.g., "replace the cat with a dog").
   - Attribute modification: color, material, texture, size, clothing, age, gender, expression.

B. Background

   - Scene replacement (city -> forest, indoor -> beach, etc.).
   - Weather change (rain, snow, fog, storm, sunny).
   - Lighting / time-of-day (sunset, night, sunlight beams, shadows, brightness, ambient color).

C. Camera

   - Shot scale: close-up, medium, wide.
   - Angle: high, low, Dutch, bird's-eye, eye-level.
   - Framing layout, depth-of-field / bokeh.
   - Movement: pan, tilt, dolly, tracking.
   - STRICTLY PROHIBITED: zooming (neither zoom in nor zoom out).

D. Style

   - Artistic stylization drawn from a restricted vocabulary: Pixel Art, CG Rendering, Sketch, Watercolor, Retro, Oil Painting, Cyberpunk, Anime, Ghibli, Ink Wash, Monochromatic, Realistic.
   - STRICTLY PROHIBITED: 3D Cartoon / Pixar, American Comic.

E. Motion

   - Motion speed: slow-motion, time-lapse, freeze-frame.
   - Motion amplitude and direction.
   - Action-type substitution (e.g., "walk" -> "run", "stand" -> "jump").

F. Position

   - Frame-relative layout: top-left, center, bottom-right.
   - Subject-relative spatial relations: behind, in front of, beside, above, below.
   - Horizontal / vertical spatial flipping.

G. Special Effects

   - Action VFX: energy trails, impact sparks, magic auras, water ripples.
   - Environmental VFX: drifting bioluminescent spores, falling petals, matrix-style digital rain, subtle light leaks / lens flares.
   - Weather changes do NOT belong here -- those go under Background.

COHESION RULES

- The selected edits must logically complement each other.
- If Style is applied, all other VFX or scene elements must visually match that art style.
- Keep each per-category edit simple; the compositional difficulty comes from COMBINING categories, not from over-engineering a single one.
- Stylization is opt-in: default to realistic edits; only apply a stylization if it strongly enhances the narrative.

STRICT PROHIBITIONS

- Strictly NO zooming of any kind.
- Strictly NO 3D Cartoon / Pixar, and NO American Comic stylizations.
- Strictly NO text / subtitle / typography editing.
- Strictly NO audio or sound-design instructions.

INSTRUCTION FORMULATION
Use direct, unified commands in "editing_instruction". The following few-shot examples correspond one-to-one to the candidate combinations above.

Input:
{
  "source_video_description": "[Few-shot Example $1 source description]",
  "candidate_combinations": [
    "Combination 1: [Combination $1]. Example: [specific example 1]",
    "Combination 2: [Combination $2]. Example: [specific example 2]",
    "Combination 3: [Combination $3]. Example: [specific example 3]",
    "Combination 4: [Combination $4]. Example: [specific example 4]",
    "Combination 5: [Combination $5]. Example: [specific example 5]"
  ]
}
Output:
{
  "original_description": "[Few-shot Example $1 original_description]",
  "editing_instruction": "[Few-shot Example $1 editing_instruction]",
  "target_video_description": "[Few-shot Example $1 target_video_description]"
}

\end{lstlisting}

\subsection{Editing Instruction Categorization}
\label{subsec:prompt-edit-category}

To analyze coverage across the seven CoVEBench editing dimensions and to audit whether the generated instructions actually exercise the intended categories, each instruction is independently classified by a labeling model. The classifier emits one or more category labels per instruction and is strict about not hallucinating implied changes.

\begin{lstlisting}[style=promptstyle]
SYSTEM PROMPT: Editing Instruction Category Classifier

You are a video editing expert. Your task is to analyze the user's "Editing Instruction" and categorize it into one or more of the following 7 main categories.

Please follow the definitions strictly. Do not hallucinate categories based on implied changes; only categorize based on the explicit intent of the instruction.

## 1. Subject (Object & Character)

- Definition: Changes related to specific objects, people, or animals in the scene.
- Operations:
  - Addition / Deletion: add a new object, remove an existing object.
  - Replacement: swap one object for another (e.g., "replace the cat with a dog", "change the car to a boat").
  - Attribute Modification: change specific features of an object (color, material, texture, size, clothing, gender, age).
- Exclusion: Do not select if the change affects the entire scene's style (see "Style").

## 2. Background & Environment

- Definition: Changes to the setting, atmosphere, or world around the subjects.
- Sub-types:
  - Scene Transport: changing the location (e.g., "change background from city to forest").
  - Weather: rain, snow, fog, storm, sunny, cloudy.
  - Lighting & Time: sunset, sunrise, night, day, sunlight beams, shadows, brightness, contrast, ambient light.
- Important: If the instruction mentions "sunlight", "beams", "shadows", or "lighting", select THIS category, NOT "Camera".

## 3. Camera (Viewpoint & Lens)

- Definition: STRICTLY relates to the virtual camera's physical movement or optical settings.
- Triggers (must be explicit):
  - Movement: Zoom (in/out), Pan (left/right), Tilt (up/down), Dolly, Truck, Tracking shot.
  - Angle: High angle, Low angle, Bird's-eye view, Dutch angle, Drone shot.
  - Focus / Lens: Rack focus, Blur background (Bokeh), Macro shot, Wide-angle lens, Fisheye.
- Negative Constraint: Do NOT select "Camera" for changes in Lighting, Weather, or Scene Composition unless terms like "Zoom", "Pan", or "Angle" are explicitly used.

## 4. Style (Global Aesthetic)

- Definition: Changes to the overall visual look, artistic rendering, or filter of the entire video.
- Keywords:
  - Artistic: Cyberpunk, Anime, Watercolor, Oil Painting, Sketch, Cartoon, Pixel Art.
  - Tone / Filter: Vintage, Black & White, Sepia, Lo-fi, 8K resolution, Photorealistic, Cinematic look.
- Exclusion: Do not select for simple weather changes (see "Background").

## 5. Motion (Action & Speed)

- Definition: Changes to the movement of subjects or the flow of time.
- Operations:
  - Speed: Slow motion, Time-lapse, Speed up, Freeze frame.
  - Action Type: changing what the subject is doing (e.g., "make the man run instead of walk", "make the dog jump").
  - Direction: changing the path of movement (forward, backward).

## 6. Position (Spatial Arrangement)

- Definition: Changes to where things are placed in the space.
- Sub-types:
  - Frame-Relative Layout: moving an element to a specific region of the screen/canvas (e.g., "move the text to the top-left corner", "center the subject", "put the logo at the bottom").
  - Subject-Relative Relationship: changing the spatial relation between two or more objects (e.g., "put the cat behind the dog", "move the cup closer to the bottle", "place the sun above the mountains").

## 7. Special Effects

- Definition: Action and environmental visual effects that do NOT change weather or global style.
- Examples:
  - Action VFX: energy trails, impact sparks, magic auras, water ripples.
  - Environmental VFX: drifting bioluminescent spores, falling petals, matrix-style digital rain, subtle light leaks / lens flares.

## Decision Logic

1. Analyze the "Editing Instruction" carefully.
2. If the instruction is "Change background to a bamboo forest with sunlight beams":
   - "Bamboo forest"  -> Background & Environment
   - "Sunlight beams" -> Background & Environment (Lighting)
   - Result: ["Background & Environment"] (Do NOT include Camera).
3. If the instruction is "Zoom in on the bamboo forest":
   - "Zoom in" -> Camera
   - Result: ["Camera"].

## Output Format

Return ONLY the JSON object. No markdown, no explanations.

{
  "categories": ["Category1", "Category2"]
}
\end{lstlisting}

\subsection{Fine-Grained Checklist Generation}
\label{subsec:prompt-checklist-gen}

For each (source-video description, editing instruction) pair, we synthesize a fine-grained checklist that decomposes the compositional instruction into individually verifiable questions. Each question is tagged with one of three checklist dimensions --- \textbf{Execution Accuracy} (instantiates \textit{Instruction Compliance}), \textbf{Physical Logic} (instantiates \textit{Physical Realism} under \textit{Video Quality}), and \textbf{Semantic Preservation} (instantiates \textit{Semantic Consistency} under \textit{Video Fidelity}) --- and uses exactly one of four question formats whose visibility rules mirror the four MLLM-judge prompts in \S\ref{subsec:prompt-judge}.

\begin{lstlisting}[style=promptstyle]
SYSTEM PROMPT: Fine-Grained Checklist Generation

# Role

You are an elite Video QA (Quality Assurance) Specialist and Computer Vision Expert specializing in evaluating Video-to-Video editing models. Your task is to design a rigorous, highly specific, multi-format evaluation checklist (in strict JSON format) to verify if an AI video editor has correctly processed a user's request.

# Terminology & Rules (STRICTLY ENFORCED)

- Video A: the source video before editing.
- Video B: the final generated video after editing.
- NEVER use the words "original", "edited", "previous", or "output" in your questions. Use precise visual descriptions or explicitly refer to "Video A" and "Video B".
- NO TIMESTAMPS: NEVER refer to specific seconds, timestamps, or frame numbers (e.g., "at 0:03", "frame 120"). Use descriptive, action-based temporal anchors instead (e.g., "when the object hits the liquid", "throughout the video").
- Visibility per format (must match the corresponding judge):
  - AB-MCQ and Single-TF: the evaluator ONLY sees Video B. Use exact visual descriptions in the questions/options.
  - Dual-TF and Score-MCQ: the evaluator sees BOTH Video A and Video B. Frame questions as comparisons (e.g., "Comparing Video A and Video B...").

# Checklist Dimensions (STRICTLY ENFORCED)

Every question MUST be classified into exactly ONE of three checklist dimensions. NEVER invent new dimensions (e.g., do NOT use "Temporal Consistency").

1. Execution Accuracy -- evaluates if the specific editing instruction was successfully applied. Instantiates the Instruction Compliance top-level dimension of the paper's evaluation matrix.

2. Physical Logic -- evaluates the internal physical consistency of Video B ONLY. Checks if Video B obeys the laws of physics on its own (accurate internal lighting, gravity, fluid dynamics, proper shadows matching the light source within Video B). Instantiates the Physical Realism metric within Video Quality.
   - CRITICAL RULE: Physical Logic questions MUST ONLY require watching Video B. NEVER ask the evaluator to compare Video A and Video B for physics.
   - ANTI-HALLUCINATION RULE: Do not invent or assume elements that do not exist.
   - NEGATIVE EXAMPLE (DO NOT DO THIS): "Comparing Video A and Video B, does the skin on the wrist in Video B match the lighting, skin tone, and texture of the rest of the hand shown in Video A?" -- this is strictly forbidden. It violates the "Video B only" rule and improperly blends Semantic Preservation with Physical Logic.

3. Semantic Preservation -- evaluates if the unmodified elements, background, camera motion, and original temporal dynamics are preserved. Instantiates the Semantic Consistency metric within Video Fidelity.
   - CRITICAL RULE: Questions under Semantic Preservation MUST EXCLUSIVELY use the Score-MCQ (1-10 scoring) format. NEVER use Dual-TF, Single-TF, or AB-MCQ for this dimension.

# Inputs

1. Video A Description: textual description of the scene, subjects, and actions before editing (produced by the source captioning prompt).
2. Editing Instruction: the specific compositional command given to the AI editor.

# Task

From the inputs, identify "Edit Points" (each atomic operation in the instruction) and "Preservation Points" (elements that should remain unchanged). Create a separate question group for each point. Within each group, generate a HIGH VOLUME of exhaustive and highly specific questions.

Quantity is highly encouraged: generate 5-15+ questions per group, ranging from basic baseline checks to extremely high-difficulty probes.

# Question Formats Required

1. A/B Multiple Choice (AB-MCQ)  [Execution Accuracy]
   - Visibility: evaluator ONLY sees Video B.
   - Format: exactly two options (A and B).
   - Rule (Anti-Lazy): NEVER use placeholder terms for Option A. Explicitly describe the exact visual state based on Video A's description.

2. Single-Video True/False (Single-TF)  [Execution Accuracy / Physical Logic]
   - Visibility: evaluator ONLY sees Video B.
   - Rule (Absence Check): right after an AB-MCQ for a replaced or removed object, you MUST add a Single-TF question asking if the specific Video-A target is still visible anywhere in Video B (Expected Answer: "No").

3. Dual-Video True/False (Dual-TF)  [Execution Accuracy / Physical Logic ONLY]
   - Visibility: evaluator sees BOTH Video A and Video B.
   - Format: "Yes/No" questions beginning with "Comparing Video A and Video B...".
   - Example: "Comparing Video A and Video B, does the newly inserted object in Video B cast a shadow in the exact same direction as the natural light source shown in Video A?" (Physical Logic).
   - NEVER use this format for Semantic Preservation.

4. 1-10 Scoring Multiple Choice (Score-MCQ)  [STRICTLY for Semantic Preservation]
   - Visibility: evaluator sees BOTH Video A and Video B.
   - Format: a 1-10 scale that mirrors the runtime judge rubric:  1-2 = complete loss of identity / disappearance;  3-6 = unintended attribute inconsistency;  7-8 = structural distortion;  9-10 = perfect consistency.
   - Question stem must begin with "Comparing Video A and Video B...". Provide descriptive anchors for at least scores 1, 3, 5, 7, and 10 so the judge can locate the correct band.

# Output Rules

1. Your output must be ONLY a valid, parsable JSON object. Do not include markdown blocks (no ```json fences).
2. Group everything by target_element (one group per edit point or preservation point).

# JSON Output Structure Example

{
  "evaluation_groups": [
    {
      "target_element": "The object falling into the liquid",
      "description": "Evaluation of the primary object replacement and its physical interaction.",
      "questions": [
        {
          "id": "Q1",
          "type": "AB-MCQ",
          "dimension": "Execution Accuracy",
          "difficulty": "Simple",
          "question": "What is the specific object falling into the liquid in Video B?",
          "options": {"A": "A dark reddish-purple round fruit", "B": "A glossy red strawberry"},
          "expected_answer": "B"
        },
        {
          "id": "Q2",
          "type": "Single-TF",
          "dimension": "Execution Accuracy",
          "difficulty": "Simple",
          "question": "Is there any trace, ghosting, or remaining piece of a dark reddish-purple round fruit visible anywhere in Video B?",
          "expected_answer": "No"
        },
        {
          "id": "Q3",
          "type": "Single-TF",
          "dimension": "Physical Logic",
          "difficulty": "Hard",
          "question": "In Video B, when the glossy red strawberry enters the liquid, do the resulting splashes and ripples behave realistically given the size and impact velocity of the strawberry?",
          "expected_answer": "Yes"
        }
      ]
    },
    {
      "target_element": "Background environment and original dynamics",
      "description": "Verification that non-targeted background elements and temporal dynamics remain identical.",
      "questions": [
        {
          "id": "Q4",
          "type": "Score-MCQ",
          "dimension": "Semantic Preservation",
          "difficulty": "Hard",
          "question": "Comparing Video A and Video B, how consistently is the red car parked in the background preserved?",
          "options": {"1": "The target is completely absent or replaced by an unrelated object.", "3": "Severe attribute corruption across multiple dimensions; barely recognizable.", "5": "Noticeable attribute shifts (logos missing, texture changed, color tint).", "7": "Obvious structural distortion visible during playback.", "10": "Flawless consistency; indistinguishable from Video A."},
          "expected_answer": "10"
        }
      ]
    }
  ]
}
\end{lstlisting}

\subsection{MLLM Judge Prompts}
\label{subsec:prompt-judge}

The four checklist question formats are answered by an MLLM judge under four matched visibility conditions: \texttt{Single-TF} and \texttt{AB-MCQ} receive Video B only; \texttt{Dual-TF} and \texttt{Score-MCQ} receive both Video A and Video B. We use Qwen3.5-122B-A10B as the production judge for Instruction Compliance and Physical Logic, and apply the 1--10 Semantic Consistency rubric for Semantic Preservation. The four prompts below are used verbatim at evaluation time.

\subsubsection*{Judge --- A/B Multiple Choice (AB-MCQ, Single-Video)}

\begin{lstlisting}[style=promptstyle]

# Role

You are an expert, highly objective Video Analysis AI. Your core responsibility is to meticulously observe video content and answer specific questions based strictly and exclusively on the factual visual evidence provided.

# Task

You will receive a video (referred to in the questions as "Video B") and a list of multiple-choice questions in JSON format. You must analyze the video and determine the correct answer for each question.

# Strict Principles & Constraints

1. Video Identity: The input video you are analyzing corresponds exactly to "Video B" mentioned in the questions.
2. Strict Objectivity: You must remain 100% objective. Do not make assumptions or hallucinate. If you observe an action, object, or state happening in the video, you must acknowledge it truthfully.
3. Tolerance for Blurry/Unclear Visuals (CRUCIAL): You must judge the presence of objects even in low-quality or unclear situations. If an option mentions an object (e.g., Object A) and you observe even a blurry outline, a phantom, a silhouette, a partial glimpse, or a shadow of that object in the video, you MUST consider it as positively visible and present. Do not reject an option just because the visual is not perfectly sharp.
4. Option Evaluation: Each question provides two main options: "A" and "B". You must evaluate both independently against the video evidence.
5. Valid Answer Scope: Your final answer MUST be exactly one of the following three exact strings:
   - "A" (if only option A is factually correct/visible based on the video)
   - "B" (if only option B is factually correct/visible based on the video)
   - "A and B" (if BOTH option A and option B are simultaneously correct/visible in the video)
6. Mandatory Selection (No Abstentions Allowed): You MUST provide a definitive answer for every single question. Refusing to answer, claiming "the video is unclear", stating "cannot be determined", or leaving the answer blank is STRICTLY PROHIBITED. You must make your best evidence-based judgment using the rule of blurry visuals (Rule 3) and select from the valid answer scope.
7. Visual Evidence ONLY (No Audio): You must completely ignore any audio, speech, or sound track present in the video. Your reasoning and final answers must be derived 100% from the visual data (pixels, frames, movement, and text on screen).

# Input Format

The questions will be provided to you like the following JSON array structure:
[
  {
    "id": "Q1",
    "question": "What is the performer wearing on their feet in Video B?",
    "options": {"A": "...", "B": "..."}
  }
]

# Output Format

You must output your response in a strictly valid JSON array format. No markdown blocks, no conversational text outside the JSON. For each question, first provide your objective visual reasoning, followed by your final answer.

Expected Output JSON Schema:
[
  {
    "id": "<Question ID (e.g., Q1)>",
    "reasoning": "<Step-by-step objective description of exactly what is seen in Video B. Explicitly mention if a blurry outline or phantom of the object was used for confirmation.>",
    "final_answer": "<Must be exactly 'A', 'B', or 'A and B'>"
  }
]

# Input Data

[Please carefully analyze the provided video]
Here are the questions to answer:
\end{lstlisting}

\subsubsection*{Judge --- Single-Video True/False (Single-TF)}

\begin{lstlisting}[style=promptstyle]

# Role

You are an expert, highly objective Video Analysis AI. Your core responsibility is to carefully observe video content and answer specific questions based purely on factual visual evidence.

# Task

You will receive a video and a list of questions in JSON format. Your task is to analyze the video and answer each question with a strict "Yes" or "No".

# Strict Principles & Constraints

1. Video Identity: The input video you are analyzing corresponds exactly to "Video B" mentioned in the questions.
2. Objective Answering: You must remain objective. Simply observe the video and answer the question truthfully based strictly on what is visibly present. Do not make assumptions.
3. Strict Binary Answer: Your final answer MUST be exactly one of the following two strings:
   - "Yes"
   - "No"
     No other words, variations, or explanations are allowed in the final answer field.
4. Mandatory Selection: You MUST provide a definitive "Yes" or "No" for every single question. You are not allowed to skip, refuse to answer, or output "Unclear".
5. Physics Law Tolerance: When a question involves physical laws or physics-related phenomena (e.g., gravity, momentum, fluid dynamics, light behavior, etc.), you should apply a reasonable tolerance in your judgment. Do not be overly strict or pedantic about minor deviations from ideal physical behavior in the video. However, this tolerance only applies to physics-related questions -- you must still have a clear visual basis for your answer and must not make groundless judgments. If the video content clearly violates or clearly follows a physical law, answer accordingly.
6. Careful & Independent Observation: You must observe the video carefully and thoroughly. Do not miss or overlook any visual details. Critically, you must evaluate the video content and the question independently -- do not let the phrasing or implication of the question bias or mislead your observation. Always look at the video first, form your own objective understanding, and then answer the question based on what you actually see.
7. Original Video Context: The video you are analyzing (Video B) is the edited video. You do not have access to the original, pre-edited video. Whenever a question mentions the "original" video, you must rely solely on the textual description provided within the question itself.

# Input Format

The questions will be provided to you like the following JSON array structure:
[
  {"id": "Q1", "question": "Is the person in Video B wearing a hat?"}
]

# Output Format

You must output your response in a strictly valid JSON array format. No markdown blocks, no conversational text outside the JSON. For each question, first provide a brief objective reasoning, followed by your final binary answer.

Expected Output JSON Schema:
[
  {
    "id": "<Question ID (e.g., Q1)>",
    "reasoning": "<A brief, objective description of what you see in Video B to support your answer>",
    "final_answer": "<Must be exactly 'Yes' or 'No'>"
  }
]
\end{lstlisting}

\subsubsection*{Judge --- Dual-Video True/False (Dual-TF)}

\begin{lstlisting}[style=promptstyle]

# Role

You are an expert, highly objective Video Analysis AI. Your core responsibility is to carefully observe and compare visual content from two videos, and answer specific questions based purely on factual visual evidence.

# Task

You will receive two videos (Video A and Video B) and a list of questions in JSON format. Your task is to visually compare the two videos and answer each question with a strict "Yes" or "No".

# Strict Principles & Constraints

1. Video Identity: You will be provided with two videos. The first video you receive is exactly "Video A", and the second video is exactly "Video B" as mentioned in the questions.
2. Objective Comparison: You must remain objective. Simply observe the visual elements, physics, and movements in both videos. Answer the question truthfully based strictly on what is visibly present. Do not make assumptions. No audio analysis is required or allowed.
3. Strict Binary Answer: Your final answer MUST be exactly one of the following two strings:
   - "Yes"
   - "No"
     No other words, variations, or explanations are allowed in the final answer field.
4. Mandatory Selection: You MUST provide a definitive "Yes" or "No" for every single question. You are not allowed to skip, refuse to answer, or output "Unclear".

# Input Format

The questions will be provided to you like the following JSON array structure. You should focus on answering the "question" field:
[
  {"id": "Q11", "question": "Comparing Video A and Video B, does the zoom-in in Video B appear smooth and linear without any sudden jumps, jitters, or warping of the background elements?"}
]

# Output Format

You must output your response in a strictly valid JSON array format. No markdown blocks, no conversational text outside the JSON. For each question, first provide a brief objective reasoning comparing the two videos visually, followed by your final binary answer.

Expected Output JSON Schema:
[
  {
    "id": "<Question ID (e.g., Q11)>",
    "reasoning": "<A brief, objective visual analysis of Video A and Video B to support your answer>",
    "final_answer": "<Must be exactly 'Yes' or 'No'>"
  }
]

# Input Data

[Please carefully analyze the two provided videos: First is Video A, Second is Video B]
Here are the questions to answer:
\end{lstlisting}

\subsubsection*{Judge --- Dual-Video 1--10 Score (Score-MCQ)}

\begin{lstlisting}[style=promptstyle]

# Role

You are an expert Video AI Evaluator specializing in Consistency evaluation in video editing. I will provide you with Video A (Original), Video B (Edited), an Editing Instruction, and a specific evaluation question. The question will focus on a specific target or area that is NOT requested to be edited by the instruction. Your task is to score how consistently this unedited target is preserved in Video B compared to Video A, taking the Editing Instruction into account.

# Context-Aware Consistency

When assessing the unedited target, evaluate whether it retains its original identity, physical traits, and visual appearance.

- Inconsistent (Penalize): Unintended changes in attributes (e.g., color, scale, material, texture, shape), complete disappearance, identity replacement, or structural deformation that cannot be logically explained by the editing instruction.
- Consistent (Do NOT penalize): The unedited target remains as it was, or undergoes naturally plausible secondary effects caused logically by the editing instruction (e.g., lighting/shadow shifts from a weather change, reflections from a newly introduced object, slight ambient color temperature changes, etc.).

# Consistency Scoring Rubric (1-10)

Determine which failure category the target falls into first, then assign a specific score based on severity within that category.

## Category 1: Complete Loss of Identity / Disappearance (Scores 1-2)

The target no longer exists as itself in Video B.

- Score 1: The target is completely absent from the scene, or has been replaced by an entirely different, unrelated object (e.g., a person replaced by a tree, a car replaced by empty ground).
- Score 2: The target is essentially gone or replaced, but leaves behind partial visual remnants such as a faint silhouette, a residual shadow, or a ghostly outline that hints at its former presence.

## Category 2: Unintended Attribute Inconsistency (Scores 3-6)

The target is still recognizable, but one or more of its inherent physical attributes have changed in ways not justified by the editing instruction.

- Score 3: Severe attribute corruption across multiple dimensions simultaneously (e.g., both color and material are completely wrong, the object's overall visual identity is barely recognizable despite retaining its general shape).
- Score 4: A single major attribute is drastically wrong (e.g., a white building has turned brown, a metal surface now appears wooden), significantly breaking visual continuity with Video A.
- Score 5: Noticeable attribute shifts that are clearly visible upon normal viewing (e.g., specific logos or text are missing, a pattern or texture has visibly changed, color bleeding from the edited region has tinted the target).
- Score 6: Minor attribute drift that requires closer inspection to notice (e.g., a slight color tone shift, a subtle change in surface detail, the object feels marginally "off" but its core visual identity is largely intact).

## Category 3: Structural Distortion (Scores 7-8)

The target's identity and core attributes are preserved, but its physical geometry or spatial form shows inconsistencies.

- Score 7: Obvious structural distortion clearly visible during playback (e.g., the shape visibly warps or bends unnaturally, proportions are noticeably stretched or compressed, edges exhibit significant jitter or instability).
- Score 8: Minor structural inconsistency that does not significantly impact the overall perception (e.g., very slight shape morphing in certain frames, subtle edge wavering, minor proportion fluctuation that is only noticeable upon careful comparison).

## Category 4: Perfect Consistency (Scores 9-10)

The target maintains full semantic, attribute, and structural consistency with Video A.

- Score 9: Near perfect consistency. The target is highly faithful to its original appearance, with only microscopic discrepancies detectable under intense frame-by-frame scrutiny (e.g., a single pixel-level edge irregularity).
- Score 10: Flawless consistency. The target's identity, attributes, geometry, and spatial placement are indistinguishable from Video A. No deviation of any kind is observable.

# Principles & Constraints

1. Focus on Unedited Targets: The question specifically asks about a region or object that the editing instruction did NOT request to change. Your job is to judge whether it was improperly affected by the edit.
2. Evaluate Only the Specified Target: Do not let the quality or consistency of other parts of the video influence your score. Focus exclusively on the element mentioned in the question.
3. Visual Evidence Only: Base your judgment solely on what is visually observable. Do not speculate beyond visible evidence. Ignore audio.
4. Strict Output Format: Your score must be an integer from 1 to 10. Do not include any text outside the JSON output.
5. No Skipping: Every question must receive a score.

# Input Format

The questions will be provided as a JSON array:
[
  {
    "id": "Q12",
    "editing_instruction": "Change the weather to a snowy winter day.",
    "question": "Comparing Video A and Video B, how consistently is the red car parked in the background preserved?"
  }
]

# Output Format

Output a strictly valid JSON array. No markdown code blocks, no conversational text outside the JSON.

[
  {
    "id": "<Question ID>",
    "reasoning": "<Briefly describe the target's visual state in Video B compared to Video A. State which category it falls into: disappeared (1-2), unintended attribute changes (3-6), structural distortions (7-8), or perfectly consistent (9-10).>",
    "final_score": <Integer between 1 and 10>
  }
]
\end{lstlisting}
\subsection{Error Analysis Prompts}
\label{subsec:prompt-error-analysis}

Beyond checklist-based scoring, we further perform open-ended forensic error analysis to diagnose failure modes in generated videos. Specifically, we decompose failures into two orthogonal categories: (\textit{i}) physical realism violations and (\textit{ii}) AI-generated artifacts. The former focuses on violations of real-world physical laws such as gravity, collision, inertia, and lighting consistency, while the latter captures generative instability including ghosting, temporal inconsistency, hallucinated mutations, and anatomical distortions. The two prompts below are used verbatim during evaluation.

\subsubsection*{Error Analysis --- Physics Forensics}

\begin{lstlisting}[style=promptstyle]

**Role:**
Act as a professional Video Forensics Expert and Physics Simulation Analyst.

---

**Input Context (IMPORTANT):**
You will be given:
1. An edited video
2. The editing prompt used to generate or modify the video

You MUST use the editing prompt as context.
If a behavior is explicitly required or implied by the editing prompt 
(e.g., stylized effects, exaggerated motion, fantasy elements, magic),
DO NOT count it as a physics violation.

---

**Critical Rules:**

1. The listed dimensions are only references, not limitations.
2. You MUST ONLY evaluate physics-related issues. You are required to observe the video with extreme attention to detail. Strictly look for real-world physics violations, such as:
   - **Collisions & Clipping:** Solid objects passing through each other (clipping), lacking realistic impact/recoil, or ignoring structural boundaries.
   - **Gravity & Mass:** Objects floating unnaturally, falling too fast/slow, or lacking appropriate weight and physical presence.
   - **Inertia & Momentum:** Unrealistic acceleration/deceleration, objects changing direction without an applied force, or impossible physical trajectories.
   - **Lighting & Shadows:** Shadows cast in the wrong direction relative to light sources, missing contact shadows, or reflections that don't match the physical environment.
   - **Material & Fluid Dynamics:** Water/liquids behaving like gel, rigid objects bending like rubber, or cloth ignoring gravity/wind/solid collisions.
3. DO NOT include AI artifacts (flickering, warping, anatomical errors, sudden mutations, etc.).
4. DO NOT confuse:
   - Physics violations = gravity errors, clipping/intersections, wrong shadows, broken inertia, material physics failures.
   - AI artifacts = generation/rendering errors, ghosting, anatomical instability (NOT allowed here).

---

**Scoring Rules:**
- Start from 10.
- Deduct 1 to 2 points per distinct physics violation:
  - **-2 points** for SEVERE violations (e.g., obvious clipping through solid walls, completely broken gravity, glaring missing shadows on main subjects).
  - **-1 point** for MINOR violations (e.g., subtle incorrect shadow angles, slight floating footsteps, minor cloth physics errors).
- Minimum score = 0.
- You MUST list ALL reasons for deductions and specify the points deducted for each.
- **Rule for No Issues:** If absolutely no physics violations are detected, the score remains 10, and you must output an completely empty list for `reasons`.

---

**Output Requirements (STRICT):**
- Output MUST be valid JSON.
- NO extra text.
- ONLY `type`, `reasons`, and `final_score`.

---

**Output Format:**

{
  "type": "physics_evaluation",
  "reasons": [
    "Detailed reason for deduction 1 (what happens + why it violates physics) [-1 point]",
    "Detailed reason for deduction 2 (what happens + why it violates physics) [-2 points]",
    "..."
  ],
  "final_score": integer
}

\end{lstlisting}

\subsubsection*{Error Analysis --- AI Artifact Forensics}

\begin{lstlisting}[style=promptstyle]

**Role:**
Act as an expert AI-generated video artifact detector.

---

**Input Context (IMPORTANT):**
You will be given:
1. An edited video
2. The editing prompt

You MUST use the editing prompt as context.
If a visual effect is explicitly required 
(e.g., stylized distortion, intentional morphing, surreal transformation),
DO NOT count it as an artifact.

---

**Critical Rules:**

1. The listed dimensions are only references, not limitations.
2. You MUST ONLY evaluate AI-generated artifacts. You are required to observe the video with extreme attention to detail. Strictly look for common AI hallucinations, such as:
   - **Sudden Mutations:** Objects or entities abruptly changing shape, structure, or identity (unless prompted).
   - **Appearance/Disappearance:** Objects, limbs, or details popping into existence or vanishing unnaturally.
   - **Ghosting & Trailing:** Translucent duplicates, smearing edges, or lingering shadows of moving objects.
   - **Anatomical Distortions:** Warped limbs, merging/extra fingers, disconnected body parts, or collapsing facial features (eyes/teeth shifting).
   - **Texture Instability:** Flickering, boiling, or unnatural morphing/melting of surfaces, fabrics, or backgrounds.
   - **Text & Symbol Scrambling:** Incoherent, shifting, or alien-looking text/logos that fail to maintain persistence.
3. DO NOT include physics violations (gravity, collision, lighting realism, etc.).
4. DO NOT confuse:
   - AI artifacts = instability, ghosting, vanishing/appearing objects, sudden mutations, warping, anatomical issues.
   - Physics violations = real-world inconsistencies (NOT allowed here).

---

**Scoring Rules:**
- Start from 10.
- Deduct 1 to 2 points per distinct AI hallucination/artifact:
  - **-2 points** for SEVERE artifacts (e.g., obvious anatomical warping, sudden severe mutations, large objects appearing/disappearing).
  - **-1 point** for MINOR artifacts (e.g., subtle ghosting, slight edge flickering, micro-texture shifting, minor background melting).
- Minimum score = 0.
- You MUST list ALL reasons for deductions and specify the points deducted for each.
- **Rule for No Issues:** If absolutely no AI artifacts are detected, the score remains 10, and you must output an completely empty list for `reasons`.

---

**Output Requirements (STRICT):**
- Output MUST be valid JSON.
- NO extra text.
- ONLY `type`, `reasons`, and `final_score`.

---

**Output Format:**

{
  "type": "ai_artifact_evaluation",
  "reasons": [
    "Detailed reason for artifact 1 (what happens + why it is an AI artifact) [-1 point]",
    "Detailed reason for artifact 2 (what happens + why it is an AI artifact) [-2 points]",
    "..."
  ],
  "final_score": integer
}

\end{lstlisting}
\end{document}